\documentclass[11pt]{article}

\usepackage[final]{acl}

\usepackage{times}
\usepackage{latexsym}
\usepackage[T1]{fontenc}
\usepackage[utf8]{inputenc}
\usepackage{microtype}
\usepackage{inconsolata}
\usepackage{graphicx}

\usepackage{xurl}
\usepackage{hyperref}
\usepackage{pgfplots}
\pgfplotsset{compat=1.18}
\usepackage{tikz}
\usepackage{booktabs}
\usepackage{xcolor}
\usepackage{array}
\usepackage{multirow}
\usepackage[normalem]{ulem}
\usepackage{tabularx}
\usepackage{makecell}
\usepackage{tabularray}
\usepackage{amsmath}
\usepackage{amssymb}
\usepackage{caption}
\usepackage[export]{adjustbox}
\usepackage{soul}
\sethlcolor{yellow!35}
\definecolor{revcolor}{HTML}{B45F06}

\usepackage[linesnumbered,ruled,vlined]{algorithm2e}
\SetKwInput{KwIn}{Input}
\SetKwInput{KwOut}{Output}
\SetKwInput{KwParam}{Parameter}
\SetKw{KwRet}{Return}
\usepackage{pifont}

\usepackage[most]{tcolorbox}
\newtcolorbox{promptbox}[1][]{
  enhanced,
  breakable,
  colback=black!2,
  colframe=black!100,
  boxrule=1pt,
  colbacktitle=black!100,
  coltitle=white,
  fonttitle=\bfseries,
  title=#1,
  arc=5pt,
  top=8pt, bottom=8pt, left=6pt, right=6pt,
  titlerule=0pt,
  toptitle=2pt,
  bottomtitle=2pt,
  before skip=10pt plus 2pt minus 1pt,
  after skip=10pt plus 2pt minus 1pt
}

\definecolor{red}{RGB}{180,0,0}
\definecolor{green}{RGB}{0,120,0}
\definecolor{blue}{RGB}{0,70,140}
\definecolor{orange}{RGB}{200,110,0}
\definecolor{hlRed}{HTML}{EFD1D1}
\definecolor{hlYellow}{HTML}{FFEBAE}
\definecolor{hlGreen}{HTML}{D6EADC}

\providecommand{\Description}[1]{}
\providecommand{\ccsdesc}[2][]{}
\providecommand{\received}[2][]{}
\providecommand{\setcopyright}[1]{}
\providecommand{\copyrightyear}[1]{}
\providecommand{\acmYear}[1]{}
\providecommand{\acmDOI}[1]{}
\providecommand{\acmISBN}[1]{}
\providecommand{\acmConference}[4][]{}
\providecommand{\acmBooktitle}[1]{}
\providecommand{\keywords}[1]{}

\title{FaVOR: LLM-Based Agentic Framework for Factor Mining via Empirical Validation}

\author{
Hyeonjin Kim$^{1}$\thanks{Equal contribution.},
Minseok Kim$^{1}$\footnotemark[1],
Seunghyeon Jung$^{1}$\footnotemark[1],
Sujin Pyo$^{2}$,
Huisu Jang$^{3}$,
Woojin Lee$^{1}$\thanks{Corresponding author.} \\
$^{1}$Department of Computer Science and Artificial Intelligence, Dongguk University-Seoul \\
$^{2}$Department of Industrial Engineering, Seoul National University \\
$^{3}$School of Finance, Soongsil University \\
\texttt{\{tkrhk8011, cdi6578296, alzkdpf23, wj926\}@dgu.ac.kr}, \\
\texttt{psj1103@snu.ac.kr},
\texttt{yej523@ssu.ac.kr}
}

\begin{document}
\maketitle

\begin{abstract}
Traditional finance relies on experts to hand-craft factors through a principled process grounded in economic rationale. Recent LLM-based multi-agent systems have automated this process, scaling factor mining far beyond manual effort. However, these automated approaches optimize directly for returns and rarely check whether a generated factor still expresses the economic hypothesis that motivated it. We identify this inconsistency between mathematical form and economic meaning as a structural failure mode of return-oriented automation. The resulting factors blur the line between real signals and spurious correlations and break down across regime shifts. We propose \textbf{FaVOR} (\textit{\textbf{Fa}ctor \textbf{V}alidation through \textbf{O}bservable \textbf{R}easoning}), an agentic framework that restructures factor mining around hypothesis-level evidence rather than return outcomes. In place of the standard hypothesis-to-formula leap, FaVOR enforces a three-stage \textit{consistency loop} tying mathematical form to economic rationale throughout. (1) \textit{Decomposition} splits a broad economic hypothesis into independent observable conditions. (2) \textit{Validation} checks whether each factor reflects its intended condition. (3) \textit{Integration} merges them into a composite whose structure remains interpretable. On the CSI~500 and S\&P~500 in 2025, FaVOR outperforms existing baselines while remaining effective across regimes. FaVOR shows that hypothesis-grounded factor discovery produces signals that are interpretable by construction, regime-robust, and economically faithful. The code is available at \url{https://github.com/damilab/FaVOR}.
\end{abstract}

\section{Introduction}
\label{sec:sec1}

\begin{figure*}[t]
    \centerline{\includegraphics[width=0.96\textwidth]{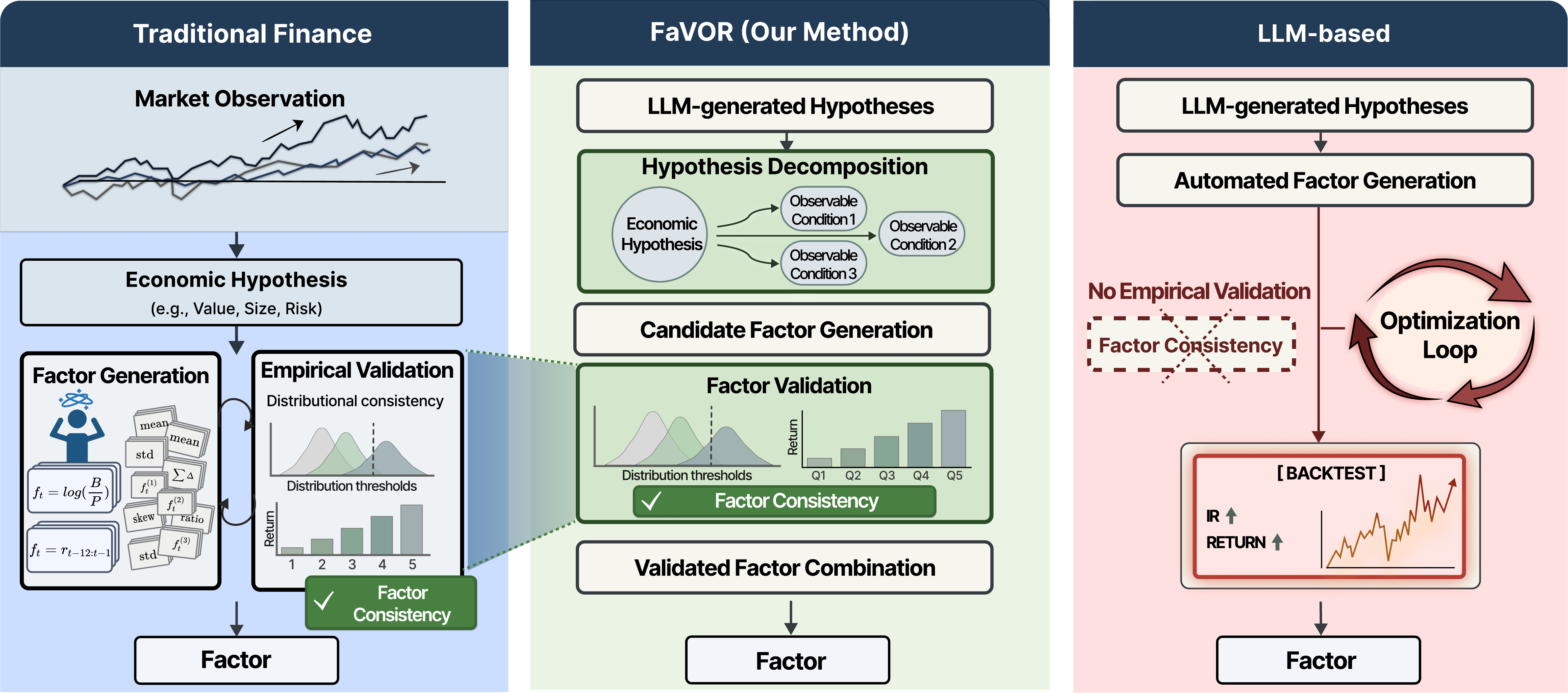}}
    \caption{Comparison of three approaches to factor discovery and validation. \textbf{Traditional} factor design relies on expert hypotheses and manual construction, supporting interpretability but limiting scalability. \textbf{LLM-based} automated mining scales generation but lacks pre-backtest validation that a factor captures its intended economic mechanism. \textbf{FaVOR} decomposes hypotheses into observable market conditions and validates candidates with distributional evidence before backtesting, combining scalability with factor-level validation.}
    \label{fig:fig1}
\end{figure*}

Large language models (LLMs) are increasingly being applied to financial analysis and investment decision-making~\cite{kim2026large}. Within this broader trend, multi-agent systems built on LLMs are reshaping quantitative investment research~\cite{lopez2025can, cheng2024empirical, zhang2024multimodal}. These systems convert qualitative investment ideas into \textit{economic hypotheses} and then translate those hypotheses into computable \textit{factors} for investment decisions~\cite{wang2025alpha, li2025r, tang2025alphaagent}. Compared with traditional factor design, LLM-based frameworks can scale factor mining beyond manual effort and report excess returns over benchmark indices~\cite{duan2025factormad, luo2025evoalpha}.

Traditional factor generation begins with market observation. Researchers first identify a \textit{market state}, an economic condition reflected in price and trading activity~\cite{karpoff1987relation, cochrane2009asset}. From these observations, they formulate an economic hypothesis that links excess returns to structural risks, market frictions, or temporary imbalances. The hypothesis becomes a factor only after rigorous empirical validation~\cite{fama1993common}. A factor in this tradition is not merely a predictive formula, but a quantitative measurement of a clearly defined economic mechanism in historical data~\cite{cochrane2011presidential, fama1996multifactor}.

Current LLM-driven approaches largely optimize factors for predictive returns, but they provide limited evidence that a factor's mathematical form reflects its intended economic mechanism~\cite{bailey2014pseudomathematics, mclean2016does, lopez2025has}. Without such validation, formulas that merely fit opaque correlations in the training window can appear indistinguishable from factors built on genuine economic mechanisms. Such formulas look strong in-sample but tend to break down when market regimes shift.

Robustness across regimes therefore requires a \textit{structurally interpretable factor} whose form maps back to the hypothesis that produced it~\cite{kou2024automate}. For LLM-based pipelines, verifying this mapping is difficult because the model returns a complete formula without exposing the intermediate economic reasoning that produced it. An explicit checking step is needed to separate factors that reflect the intended economic signal from those that overfit to noise~\cite{jiang2025alpha}.

We propose \textbf{FaVOR} (\textit{\textbf{Fa}ctor \textbf{V}alidation through \textbf{O}bservable \textbf{R}easoning}), an agentic framework that evaluates \textit{factor consistency} through a three-stage process. \textit{Decomposition} breaks a hypothesis into observable market conditions. \textit{Validation} checks whether each generated factor reflects its intended condition. \textit{Integration} identifies the combination of validated factors that best reconstructs the original hypothesis. By admitting only factors that pass all three stages, FaVOR builds structurally interpretable factors that combine empirical performance with economic rationale.

On the Chinese CSI~500 and U.S.\ S\&P~500 in the 2025 out-of-sample testing window, FaVOR delivers consistent performance across both markets. After transaction costs, the framework achieves a cumulative excess return of 0.2225 with IR of 1.5295 on the CSI~500 and 0.1123 with IR of 1.1315 on the S\&P~500. These results suggest that hypothesis-centered empirical validation can improve the economic reliability of automated factor mining.
\section{Related Work}
\label{sec:sec2}

\textbf{Traditional Foundations of Factor Investing.} Traditional quantitative research treats a factor as a measure of a specific economic mechanism rather than a formula optimized only for returns~\cite{cochrane2009asset,bender2013foundations}. This framing rests on the principle of \textit{informational efficiency}, under which latent market states are reflected in prices and trading activity~\cite{fama1970efficient}. Researchers therefore use empirical validation to test whether a market signal aligns with an explicit economic hypothesis~\cite{hou2015digesting, banz1981relationship, sloan1996stock, baker2011benchmarks}.

Prior studies show how this logic appears in factor research. \citet{basu1977investment} links valuation ratios to expected returns, while \citet{jegadeesh1993returns} studies momentum as a return pattern grounded in market behavior. Models with multiple factors extend this idea by combining observable firm characteristics and market indicators. \citet{fama1993common, fama2015five} integrate market risk, size, value, profitability, and investment patterns to explain the cross section of expected stock returns, and \citet{kakushadze2016101} shows that combining many independent alpha formulas from price, volume, and fundamentals can improve portfolio diversification and return stability. These factors remain economically interpretable, but the process depends on manual testing by domain experts and does not scale to a large candidate space.

\begin{figure*}[t]
    \centerline{\includegraphics[width=0.96\textwidth]{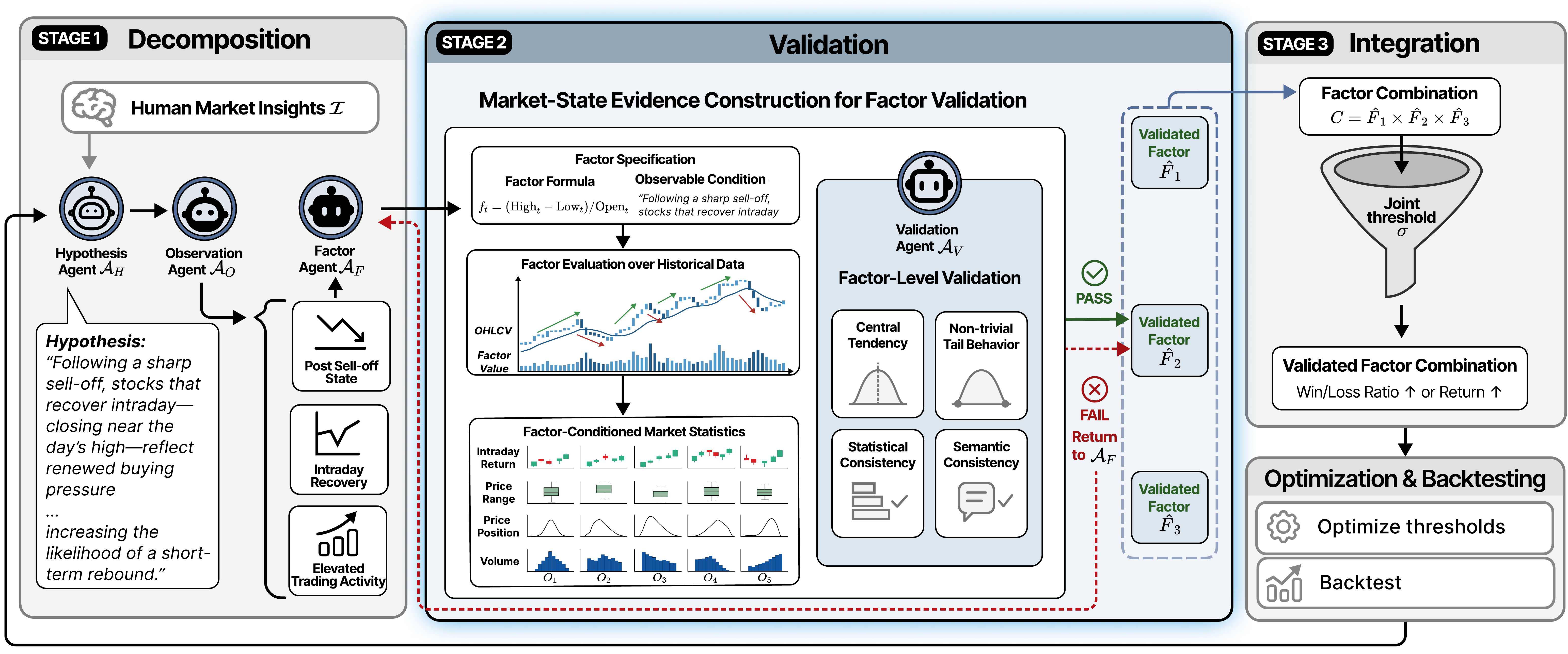}}
    \caption{Overview of the FaVOR framework: a three-stage pipeline for hypothesis-driven factor mining. Stage 1 decomposes a high-level hypothesis into observable market conditions and generates candidate factors via the Hypothesis, Observation, and Factor agents. Stage 2 validates each candidate by constructing factor-conditioned market-state statistics, with pass/fail gates that route rejected factors back to the Factor agent. Stage 3 integrates the validated factors under a joint threshold and runs threshold optimization and out-of-sample backtesting to produce stable, hypothesis-consistent trading signals.}
    \label{fig:fig2}
\end{figure*}

\textbf{Automated Factor Mining.} Recent LLM-based approaches automate factor generation by using the financial knowledge encoded in the models and their ability to enumerate candidates. FAMA~\cite{li2024can} introduces sample based selection and an experience based mechanism to reduce redundancy. \citet{cheng2024gpt} show that GPT-4 can synthesize useful factors through financial reasoning alone. Frameworks with multiple agents push the idea further by coordinating specialized agents in a collaborative search and optimization pipeline. Alpha-GPT~\cite{wang2025alpha} translates researchers' ideas into formulas and optimizes them algorithmically. RD-Agent-Quant~\cite{li2025r} jointly optimizes factors and machine learning models across the workflow from hypothesis to code. AlphaAgent~\cite{tang2025alphaagent} adds regularized exploration to promote hypothesis consistency and resist alpha decay.

Despite these advances, agents based on LLMs still evaluate candidate factors mainly through semantic fit or realized performance. Recent verification mechanisms in this line of work~\cite{wang2024quantagent, luo2025llm, lee2025your, ding2025alphaeval} provide useful checks, but natural language explanations alone do not show that a factor reflects an underlying market state. This leaves a gap between automated factor generation and the empirical validation that supports interpretability and reliability in traditional finance~\cite{kang2023deficiency, tatsat2025beyond}.
\section{Methods}
\label{sec:sec3}

We propose \textbf{FaVOR}, an agentic framework that constructs structurally interpretable factors from economic hypotheses. The framework does not evaluate a hypothesis solely by its final return. Instead, it first decomposes the hypothesis into observable market conditions, then checks whether candidate factors measure those conditions, and finally integrates the validated factors into executable trading signals. Figure~\ref{fig:fig2} summarizes this three-stage pipeline.

\subsection{Stage 1 : Hypothesis Decomposition}
\label{sec:sec3_2}

Stage 1 turns a high-level economic hypothesis into measurable components. FaVOR first generates the hypothesis from market insights, then decomposes it into observable market conditions, and finally generates candidate factors that give each condition a quantitative form.

\subsubsection{Hypothesis Generation}
\label{sec:sec3_2_1}

The hypothesis agent $\mathcal{A}_H$ takes human market insights $I$ and returns an economic hypothesis $\mathcal{A}_H(I) = h$. The hypothesis $h$ is a descriptive statement about market states and their expected impact on price movements.

\subsubsection{Hypothesis Decomposition}
\label{sec:sec3_2_2}

The hypothesis $h$ is written in natural language, so it cannot be validated directly against raw market data. The observation agent $\mathcal{A}_O$ decomposes $h$ into $m$ observable market conditions, $\mathcal{A}_O(h) = \{ o_i \mid i = 1, \dots, m \}$, where each $o_i$ captures a distinct and independent aspect of the hypothesis.

\subsubsection{Candidate Factor Generation}
\label{sec:sec3_2_3}

Given an observable market condition $o_i$, the factor agent $\mathcal{A}_F$ generates $n_i$ candidate factors, $\mathcal{A}_F(o_i) = F_i = \{ f_{i,j} \}_{j=1}^{n_i}$. Each candidate factor $f_{i,j}$ is a formula computed from market data and intended to measure $o_i$. To keep the formulas interpretable and executable, $\mathcal{A}_F$ draws operators from the predefined set in Appendix~\ref{sec:seca}. The resulting candidate factors enter Stage 2 for empirical consistency checks.

\subsection{Stage 2 : Factor-Level Validation}
\label{sec:sec3_3}

This stage evaluates the factor consistency of the candidate factors from Stage 1. Factor-level validation verifies whether each formula embodies the economic rationale of its corresponding observable market condition. It uses market variables observed at time $t$ rather than future returns at $t+T$, where $T$ denotes the holding horizon of the trading strategy. We use an LLM agent because the same statistical pattern can support one rationale while contradicting another. Consistency must therefore be judged against the hypothesis context.

\begin{figure}[ht]
    \centering
    \includegraphics[width=1.0\columnwidth]{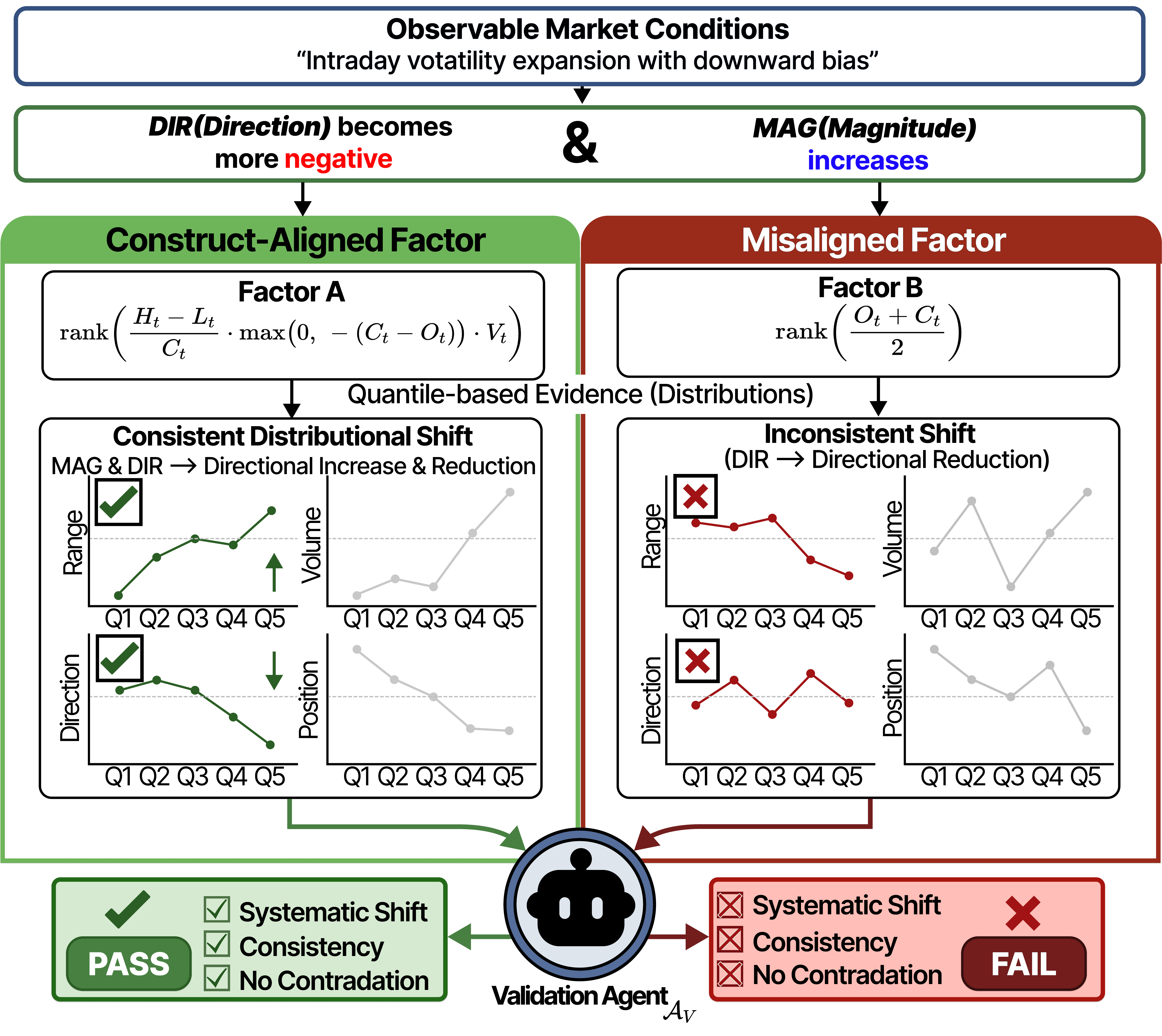}
    \caption{Illustration of factor-level validation using quantile-based distributional evidence. A construct-aligned factor (Factor A) shows systematic distributional shifts across ordered quantile bins that match the corresponding observable market condition. It therefore passes validation. In contrast, a misaligned factor (Factor B) shows unstable or contradictory patterns across bins, leading to validation failure.}
        \label{fig:fig3}
\end{figure}

\subsubsection{Historical Evidence for Consistency Evaluation}
\label{sec:sec3_3_1}

The validation agent $\mathcal{A}_V$ checks whether a candidate factor measures the observable market condition it was generated for. The goal is to test whether stronger factor values correspond to stronger evidence of the market state described in $o_i$, not whether the factor predicts future returns.

For each candidate factor, we compute factor values on the training data, sort stocks within each date, and split them into quintile bins $Q_1, \ldots, Q_5$ ordered from low to high. To describe the market state in each bin, we use four OHLCV-based variables that capture intraday return ($C - O$), daily price range ($H - L$), close position within the daily range ($(C - L)/(H - L)$), and trading volume ($V$). We denote this set by $\mathcal{X}$. For each bin and each $x \in \mathcal{X}$, we compute standard descriptive statistics (mean, median, q10, q90, and dispersion measures), and collect them into the statistical profile $S_{i,j}$ that $\mathcal{A}_V$ uses to assess factor consistency.

\subsubsection{Factor Consistency Evaluation Criteria}
\label{sec:sec3_3_2}

Given the profile $S_{i,j}$, the validation agent $\mathcal{A}_V$ decides whether the empirical patterns are consistent with the intended economic rationale $o_i$. Each check applies fixed rules to the statistics in $S_{i,j}$. The LLM is used only to map the textual observation $o_i$ to the relevant variables in $\mathcal{X}$ and their expected directions. Given this mapping, the final pass/fail decision is deterministic because it is the conjunction of four explicit rule checks.

We apply four consistency checks: (i) a systematic shift in central tendency of some $x \in \mathcal{X}$ across the ordered factor bins, (ii) non-trivial variation in the tail spread of $x$ across bins, (iii) agreement across at least two complementary statistics of the same $x$, and (iv) semantic agreement with $o_i$. The precise definitions and decision rule are deferred to Appendix~\ref{sec:secb}.

The validated factor set for each condition is $\hat{F}_i = \{ f_{i,j} \in F_i \mid \mathcal{A}_V(o_i, f_{i,j}, S_{i,j}) = \mathsf{pass} \}$. Once $\hat{F}_i$ is built for every condition, the pipeline moves to factor integration.

\subsection{Stage 3: Factor Integration}
\label{sec:sec3_4}

This stage shifts the unit of analysis from a single factor to a combination of validated factors, and asks whether such a combination realizes the original hypothesis $h$. We test this by examining how the combination's joint trading signal behaves as the entry rule becomes more selective.

\subsubsection{Candidate Combinations}
\label{sec:sec3_4_1}

For each observable condition $o_i$, let $\hat{F}_i$ denote the set of validated factors from Stage~2, which typically contains two or three formulas. We form candidate combinations by taking the Cartesian product $\mathcal{C} = \hat{F}_1 \times \hat{F}_2 \times \cdots \times \hat{F}_m$ with cardinality $|\mathcal{C}| = \prod_{i=1}^{m} |\hat{F}_i|$, and each $\mathbf{f} \in \mathcal{C}$ selects one validated factor per observable condition.

\subsubsection{Directional Selectivity}

Each combination $\mathbf{f} \in \mathcal{C}$ needs an entry rule that we can vary in selectivity, so we can test whether the direction of its realized return remains stable as the rule tightens. We call this property \textit{Directional Selectivity}. The entry rule uses a common selectivity level $\sigma \in (0, 1)$ that determines the cross-sectional extremeness required from each factor. For each $f_{i,j}$ in the combination, an observation signal triggers when $f_{i,j}$ crosses its in-sample $\sigma$-quantile in the direction implied by $o_i$ (its polarity). A trading signal $\mathrm{Signal}(\mathbf{f}, \sigma) = 1$ is triggered only when every constituent factor of $\mathbf{f}$ triggers its observation signal simultaneously, and $\mathrm{Signal}(\mathbf{f}, \sigma) = 0$ otherwise.

Increasing $\sigma$ tightens each factor's quantile cutoff and makes the joint signal more selective. It reduces the number of trading opportunities and keeps only stronger realizations of the hypothesized market state.

\subsubsection{Evaluation Criteria}

We evaluate Directional Selectivity by tracking how signal outcomes change across a ladder of selectivity levels $\sigma_1 < \sigma_2 < \cdots$. For a factor combination $\mathbf{f}$, we divide its triggered signals at each level into winning signals $\mathcal{S}_W$ with positive realized returns and losing signals $\mathcal{S}_L$ with negative realized returns. The average realized return is higher under stricter $\sigma$, confirming that selectivity targets more profitable market states. The win rate $|\mathcal{S}_W| / (|\mathcal{S}_W| + |\mathcal{S}_L|)$ is higher under stricter $\sigma$, indicating fewer false positives as the signal strengthens.

We apply these criteria at the ticker level. A ticker supports the combination only when at least one of these behaviors holds across the ladder. We then retain a factor combination only when its ticker-level support rate exceeds a fixed pass-rate threshold, and discard combinations below this threshold because stricter entry conditions do not improve the quality of their signals.

\subsection{Optimization and Backtesting}
\label{sec:sec3_5}

In Section~\ref{sec:sec3_4}, Stage~3 selects factor combinations. We turn these into executable trading signals and evaluate their realized performance. The selectivity sweep in Stage~3 is used for structural screening on the training set, while the final execution thresholds are optimized on a held-out validation set defined in Section~\ref{sec:sec4_1}. Because each factor in a combination targets a distinct observable condition, we treat its activation cutoff as independent and optimize factor-specific thresholds for each validated combination with Optuna~\cite{10.1145/3292500.3330701}.

The optimization objective is the Calmar ratio~\cite{ELING20072632}. It is defined as $\mathrm{Calmar} \;=\; \mathrm{AR} / {\lvert\mathrm{MDD}\rvert}$, where $\mathrm{AR}$ denotes the annualized return and $\mathrm{MDD}$ the maximum drawdown. This objective accounts for both return and downside risk. The selected thresholds are then fixed and backtested on the held-out test period.

Evaluation outcomes are used only in the outer-loop hypothesis memory. After each iteration, FaVOR records prior hypothesis IDs, observation descriptions, formula names, and a single scalar summary of average information ratio. The memory does not store raw returns, formula definitions, or fitted thresholds, and the pipeline code and hyperparameters are fixed before held-out evaluation. Feedback accumulates as hypothesis memory across iterations.
\section{Experiments}
\label{sec:sec4}

\begin{table*}[ht]
    \centering
    \resizebox{\textwidth}{!}{%
        \begin{tabular}{llcccccccc}
        \toprule
        & Market
        & \multicolumn{4}{c}{\textbf{CSI 500}} & \multicolumn{4}{c}{\textbf{S\&P 500}} \\
        \cmidrule(lr){3-6} \cmidrule(lr){7-10}
        Category & Method & AR & IR & MDD & CR & AR & IR & MDD & CR \\
        \midrule
        ML/DL & Linear \cite{pedregosa2011scikit}           & -0.0702  & -0.5008  & -0.1649  & -0.0778 & -0.0010 & -0.0104 & \underline{-0.1403}  & -0.0055 \\
        & XGBoost \cite{chen2016xgboost}                    & -0.0123  & -0.1642  & -0.1370  & -0.0152 & -0.0357 & -0.3345 & -0.2326  & -0.0426 \\
        & LightGBM \cite{ke2017lightgbm}                    & -0.1576  & -1.3954  & -0.2173  & -0.1530 & -0.0302 & -0.3389 & -0.2154  & -0.0353 \\
        & MLP \cite{rumelhart1986learning}                  & -0.0493  & -0.4812  & -0.1725  & -0.0537 & -0.0895 & -0.8041 & -0.1686  & -0.0956 \\
        & Transformer \cite{vaswani2017attention}           & -0.0521  & -0.3943  & -0.1760  & -0.0597 & -0.1744 & -1.5509 & -0.1860  & -0.1729 \\
        \midrule
        RL & AlphaForge \cite{shi2025alphaforge}            & -0.0133  & -0.1255  & \underline{-0.1260}  & -0.0167 & -0.0833 & -0.8617 & -0.2093  & -0.0819 \\
        & AlphaQCM \cite{zhu2025alphaqcm}                   & 0.0869   & 1.0604   & -0.1396  & 0.0803  & \underline{0.0619}  & \underline{0.8250}  & -0.2345  & 0.0583 \\
        \midrule
        LLM-based & RD-Agent \cite{li2025r}                 & \underline{0.1382}   & \underline{1.3360}   & -0.1427  & \underline{0.1452}  & 0.0604  & 0.8028  & -0.1704  & \underline{0.0624} \\
        & AlphaAgent \cite{tang2025alphaagent}              & -0.0192  & -0.1490  & -0.1369  & -0.0275 & -0.0060 & -0.0631 & -0.1428  & -0.0110 \\
        & \textbf{FaVOR}                              & \textbf{0.2067}   & \textbf{1.5295}   & \textbf{-0.0853}  & \textbf{0.2225}  & \textbf{0.1062}  & \textbf{1.1315}  & \textbf{-0.0443}  & \textbf{0.1123} \\
        \bottomrule
        \end{tabular}%
    }
    \caption{Main results on CSI 500 and S\&P 500 excess returns using train, validation, and test splits within 2022--2025. \textbf{Bold} marks the best result and \underline{underline} marks the second best.}
    \label{tab:tab1}
\end{table*}%

\begin{figure*}[ht]
    \centering
    \begin{minipage}{0.49\textwidth}
        \centering
        \includegraphics[width=\textwidth]{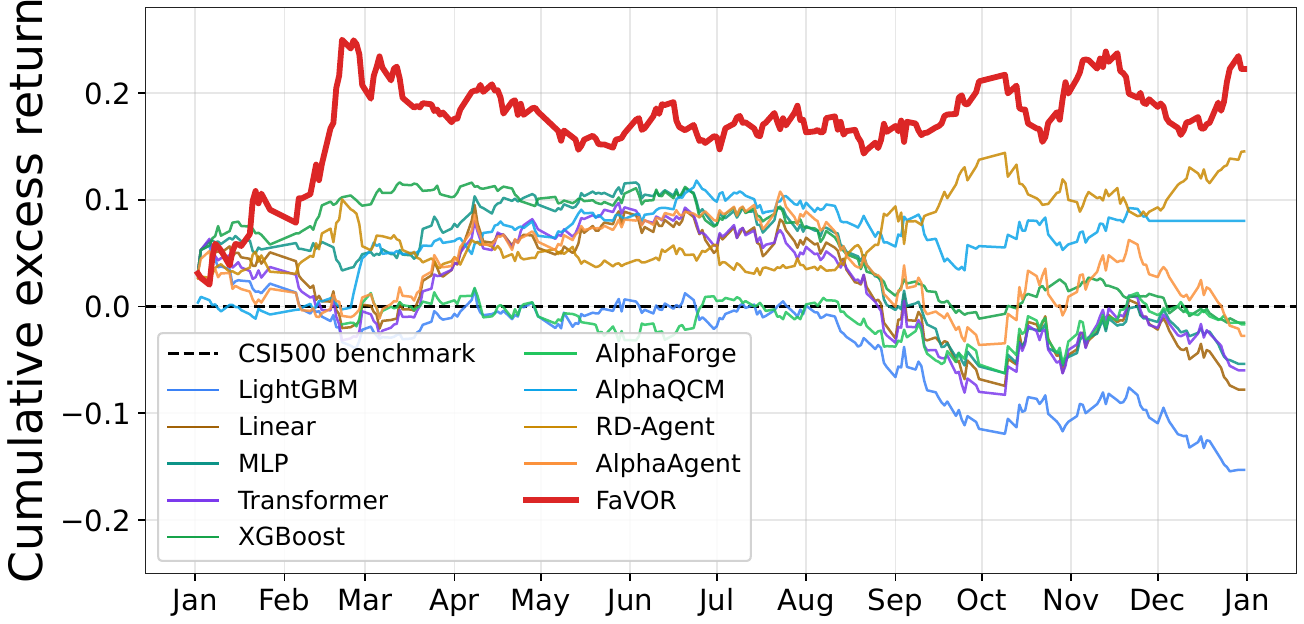}
    \end{minipage}
    \hfill
    \begin{minipage}{0.49\textwidth}
        \centering
        \includegraphics[width=\textwidth]{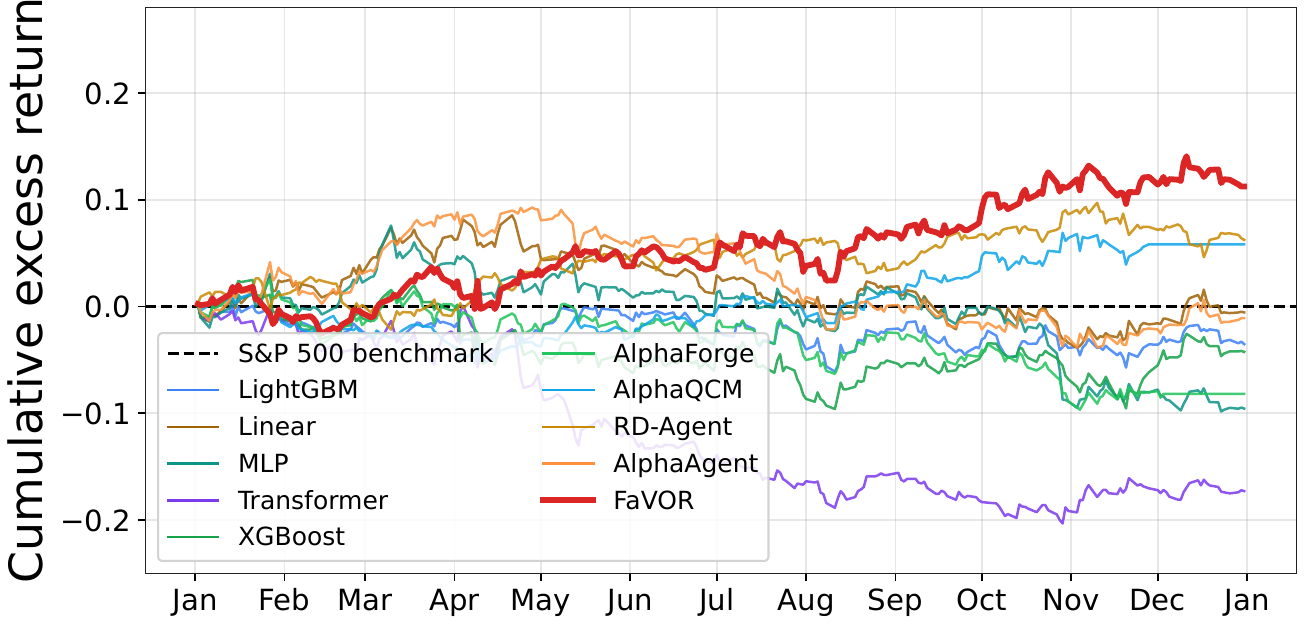}
    \end{minipage}
    \caption{Comparison of cumulative excess returns of various baselines on the CSI 500 (left) and S\&P 500 (right).}
    \label{fig:fig4}
\end{figure*}

\subsection{Experiment Settings}
\label{sec:sec4_1}

\paragraph{Metrics.} We evaluate portfolio performance with four metrics. \textbf{Annualized Return (AR)} measures the average yearly excess return over the benchmark. \textbf{Information Ratio (IR)} measures excess return scaled by tracking risk. \textbf{Cumulative Return (CR)} measures the total excess return over the full evaluation period. \textbf{Maximum Drawdown (MDD)} measures the largest decline in the portfolio's own equity curve from a previous peak. Appendix~\ref{sec:secc} provides the mathematical definitions.

\paragraph{Datasets.} We conduct backtests with Qlib~\cite{yang2020qlib} on two equity universes. The \textbf{CSI~500} covers the Chinese A-share market, and the \textbf{S\&P~500} covers the U.S.\ equity market. Raw CSI~500 data are obtained from Baostock, and S\&P~500 data are obtained from Yahoo Finance.

We split the data into training (2022-01-01 to 2023-12-31), validation (2024-01-01 to 2024-12-31), and testing (2025-01-01 to 2025-12-31) periods. Factor validation and integration are performed on the training set, threshold optimization is performed on the validation set, and the testing period is reserved exclusively for out-of-sample evaluation. Backtests use only the daily \textbf{OHLCV} fields \texttt{\$open}, \texttt{\$high}, \texttt{\$low}, \texttt{\$close}, and \texttt{\$volume}.

\paragraph{Backtest Settings.} After the trading signals and thresholds are fixed, they are applied unchanged throughout the testing period. A signal is triggered at time $t$ when the factor combination indicates that the hypothesized market state holds. Positions are opened according to the predefined execution protocol and closed according to predefined exit rules, such as the holding horizon $T$ or stop conditions. All backtest results include proportional transaction costs per traded value. For CSI~500, the cost rates are 0.0005 for buys and 0.0015 for sells. For S\&P~500, the cost rate is 0.0005 for sells only. Full hyperparameter values used across all stages are listed in Appendix~\ref{sec:secd}.

\paragraph{Agent Settings.} FaVOR uses four LLM-based agents, and we adopt \textbf{GPT-4o}~\cite{hurst2024gpt} as the default backbone for all main-table results. Decoding temperatures are set higher for the generative agents and low for the validation agent, balancing idea diversity against verdict stability. Exact values are listed in Appendix~\ref{sec:secd}. The prompt templates for all four agents are provided in Appendix~\ref{sec:sece}.

\paragraph{Baselines.} We compare FaVOR against baselines that span the paradigms most directly comparable to its hypothesis-driven, validation-centric design. (i) \textit{ML/DL methods}: \textbf{Linear}~\cite{pedregosa2011scikit}, \textbf{XGBoost}~\cite{chen2016xgboost}, \textbf{LightGBM}~\cite{ke2017lightgbm}, \textbf{MLP}~\cite{rumelhart1986learning}, and \textbf{Transformer}~\cite{vaswani2017attention}. (ii) \textit{Reinforcement Learning methods}, which optimize a return-driven reward signal under a different search formulation: \textbf{AlphaForge}~\cite{shi2025alphaforge} and \textbf{AlphaQCM}~\cite{zhu2025alphaqcm}. (iii) \textit{LLM-based methods}: \textbf{R\&D-Agent-Quant}~\cite{li2025r} and \textbf{AlphaAgent}~\cite{tang2025alphaagent}.

\begin{table*}[ht]
    \centering
    \small
    \setlength{\tabcolsep}{0.01\textwidth}
    \renewcommand{\arraystretch}{1.15}
    \begin{tabular}{@{}
    m{0.085\textwidth}
    p{0.18\textwidth}
    p{0.68\textwidth}
    @{}}
    \toprule
    
    \multicolumn{3}{@{}p{\textwidth}@{}}{\textbf{Hypothesis:} After a {\sethlcolor{hlRed}\hl{sharp sell-off}}, stocks that exhibit {\sethlcolor{hlYellow}\hl{intraday recovery, closing nearer to the daily high}}, tend to enter {\sethlcolor{hlGreen}\hl{early stabilization}}, which is associated with a short-term rebound over the next 3 trading days.} \\
    \midrule
    
    \textbf{Observable Condition} &
    \textbf{Candidate Factor} &
    \textbf{Validation reasoning based on distributional evidence} \\
    \midrule

    \multirow[t]{2}{=}{\sethlcolor{hlRed}\hl{O1: Post Sell-off State}}
    & \texttt{TS\_ZSCORE(C,10)}
    & \textbf{Reasoning:}
    "As factor intensity increases, prices close progressively lower within the daily range, indicating sustained downside pressure.
    \uline{This shift is consistently supported by aligned location and tail behavior, with DIR mean and median decreasing $(0.14 \rightarrow -0.11)$ ... and tail statistics moving downward together, characterizing a post sell-off state."}
    
    \textbf{Decision:} \textcolor{green}{\textbf{PASS}~($\checkmark$)} \\ \cmidrule(lr){2-3}
    & \texttt{TS\_ZSCORE(V, 10) * TS\_Z \linebreak SCORE(DELTA(C, 1), 10)}
    & \textbf{Reasoning:}
    "The factor does not induce a stable separation of post sell-off states.
    \uline{Upper-tail expansion $(q90 +24\text{--}157\%)$ ... coincides with sharp kurtosis compression$(-46\text{--}49\%)$, producing internally conflicting tail signals.}"
    
    \textbf{Decision:} \textcolor{red}{\textbf{FAIL}~($\times$)} \\
    \midrule

    \multirow[t]{2}{=}{\sethlcolor{hlYellow}\hl{O2: Intraday Recovery}}
    & \texttt{(close - TS\_MIN(low, 3)) / (high - TS\_MIN(low \linebreak  , 3) + eps)}
    & \textbf{Reasoning:}
    "The factor captures a clear shift in closing position within the daily range,
    with prices closing progressively nearer to the intraday high as factor strength increases.
    \uline{POS mean and median increase monotonically (0.083$\rightarrow$0.914) ...
    and the upper tail also rises (POS q90 $\rightarrow$ 1.0),}
    indicating strengthening intraday buying pressure."
    
    \textbf{Decision:} \textcolor{green}{\textbf{PASS}~($\checkmark$)} \\ \cmidrule(lr){2-3}
    & \texttt{-CORR(rank(DELTA \linebreak (logV)),} \texttt{\quad rank(DIR))}
    & \textbf{Reasoning:}
    "The induced trend does not align with the intended interpretation of factor intensity.
    \uline{Although the polarity is defined as `higher-is-more-true', the evidence decreases
    across bins ... directly contradicting the expected intraday recovery pattern.}
    As a result, the factor fails to provide a meaningful proxy for intraday recovery."
    
    \textbf{Decision:} \textcolor{red}{\textbf{FAIL}~($\times$)} \\
    \midrule

    \multirow[t]{2}{=}{\sethlcolor{hlGreen}\hl{O3: Early Stabilization}}
    & \texttt{(TS\_MIN(low, 3) - low) / (high - low + eps)}
    & \textbf{Reasoning:}
    "The factor captures a consistent narrowing of the daily range as factor strength increases.
    \uline{MAG mean and median both decrease (-21\% and -18\%) ... and tail evidence compresses
    (MAG q90 -20.5\%), with increased central concentration (kurtosis +10.9\%).}
    This pattern is consistent with early stabilization."
    
    \textbf{Decision:} \textcolor{green}{\textbf{PASS}~($\checkmark$)} \\ \cmidrule(lr){2-3}
    & \texttt{1 / (eps + TS\_STD(C, 20))}
    & \textbf{Reasoning:}
    "The factor does not provide a coherent narrowing pattern as factor strength increases.
    \uline{Tail indicators are internally inconsistent, with kurtosis decreasing (-28.4\%) ...
    while q90 shows little change (+1.8\%).}
    This inconsistency prevents a stable characterization of early stabilization."
    
    \textbf{Decision:} \textcolor{red}{\textbf{FAIL}~($\times$)} \\
    \bottomrule
    \end{tabular}
    \caption{Case study of factor-level validation under a single hypothesis. The table reports candidate factors generated from decomposed observable market conditions and their corresponding validation outcomes. For each candidate factor, we show the rationale produced by the validation agent, with underlined phrases highlighting the evidence used in factor-level validation.}
    \label{tab:tab2}
\end{table*}

\subsection{Main Results}
\label{sec:sec4_main_results}

Table~\ref{tab:tab1} reports performance on the CSI~500 and S\&P~500, evaluated on excess returns. On the CSI~500, FaVOR achieves the best result across all four metrics, reaching AR of 0.2067, IR of 1.5295, MDD of $-0.0853$, and CR of 0.2225, simultaneously improving return generation and reducing downside risk relative to every compared method.

On the S\&P~500, FaVOR again attains the best result across all four metrics, reaching AR of 0.1062, IR of 1.1315, MDD of $-0.0443$, and CR of 0.1123. This consistency suggests that FaVOR's validation step generalizes across distinct equity markets, producing effective factor combinations in both the Chinese A-share and U.S.\ universes. Appendix~\ref{sec:secf} reports the run-to-run variance over 10 independent runs and confirms that these headline numbers are reproducible.

Across the method families in Table~\ref{tab:tab1}, only FaVOR combines adaptive factor generation with an explicit distributional check between each factor and the market state it should measure, and this combination delivers the strongest return and risk profile in both markets. Appendix~\ref{sec:secg} reports the trade-level distribution and shows that gains are spread broadly across many trades.

Figure~\ref{fig:fig4} traces cumulative excess returns. On the CSI~500, FaVOR captures an early-year gain and preserves it while most baselines drift sideways and turn negative by year-end. On the S\&P~500, FaVOR remains near zero through early months and compounds steadily from mid-year onward, ending as the strongest trajectory while most baselines stay weakly positive or drift negative. Appendix~\ref{sec:sech} characterizes the 2025 market paths for both universes.

\begin{table*}[t]
    \centering
    \begingroup
    \footnotesize
    \setlength{\tabcolsep}{4pt}
    \renewcommand{\arraystretch}{1.10}

    \begin{tabular}{lcccccccc}
        \toprule
        & \multicolumn{4}{c}{\textbf{CSI 500}}
        & \multicolumn{4}{c}{\textbf{S\&P 500}} \\
        \cmidrule(lr){2-5}
        \cmidrule(lr){6-9}

        \textbf{Backbone}
        & \textbf{AR} & \textbf{IR} & \textbf{MDD} & \textbf{CR}
        & \textbf{AR} & \textbf{IR} & \textbf{MDD} & \textbf{CR} \\
        \midrule

        \textbf{GPT-4o} (default)
        & \textbf{0.2067}
        & \textbf{1.5295}
        & \underline{-0.0853}
        & \textbf{0.2225}
        & \textbf{0.1062}
        & \textbf{1.1315}
        & \textbf{-0.0443}
        & \textbf{0.1123} \\

        GPT-5.4-mini
        & \underline{0.1039}
        & \underline{1.3246}
        & \textbf{-0.0443}
        & \underline{0.1079}
        & 0.0299
        & 0.2553
        & -0.1186
        & 0.0244 \\

        Gemini-2.5-Flash
        & 0.0732
        & 0.4998
        & -0.1418
        & 0.0657
        & \underline{0.0808}
        & \underline{0.7039}
        & \underline{-0.0621}
        & \underline{0.0807} \\

        Claude-Sonnet-4.6
        & 0.0541
        & 0.5364
        & -0.0878
        & 0.0511
        & 0.0072
        & 0.0817
        & -0.0898
        & 0.0035 \\

        Llama-3.3-70B
        & 0.0449
        & 0.4183
        & -0.1313
        & 0.0407
        & 0.0146
        & 0.1545
        & -0.0801
        & 0.0107 \\

        Qwen3-235B
        & 0.0948
        & 1.0598
        & -0.0936
        & 0.0967
        & 0.0174
        & 0.1670
        & -0.1033
        & 0.0126 \\

        \bottomrule
    \end{tabular}

    \endgroup
    \caption{
        Sensitivity of FaVOR to the LLM backbone on the CSI~500 and
        S\&P~500. All pipeline settings other than the backbone are held
        fixed. The best result in each column is shown in \textbf{bold},
        and the second-best result is \underline{underlined}.
    }
    \label{tab:backbone_sensitivity}
\end{table*}

\begin{figure*}[t]
    \centering
    \begin{minipage}{0.49\textwidth}
        \centering
        \includegraphics[width=\textwidth]{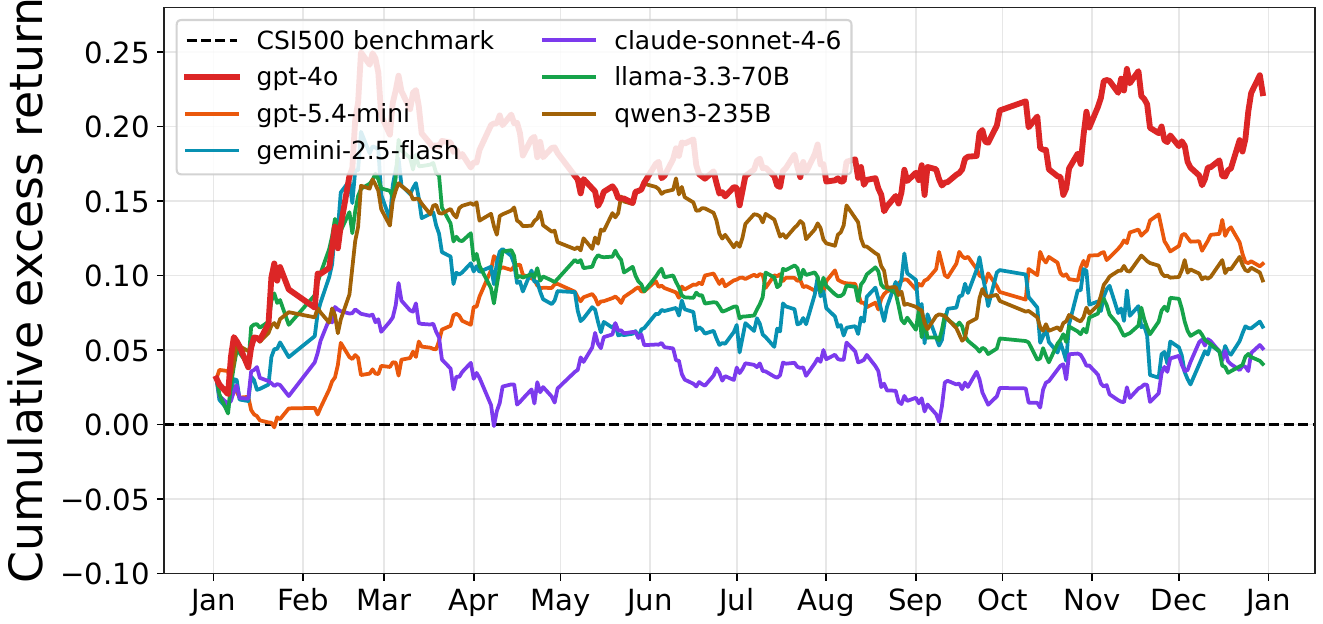}
    \end{minipage}
    \hfill
    \begin{minipage}{0.49\textwidth}
        \centering
        \includegraphics[width=\textwidth]{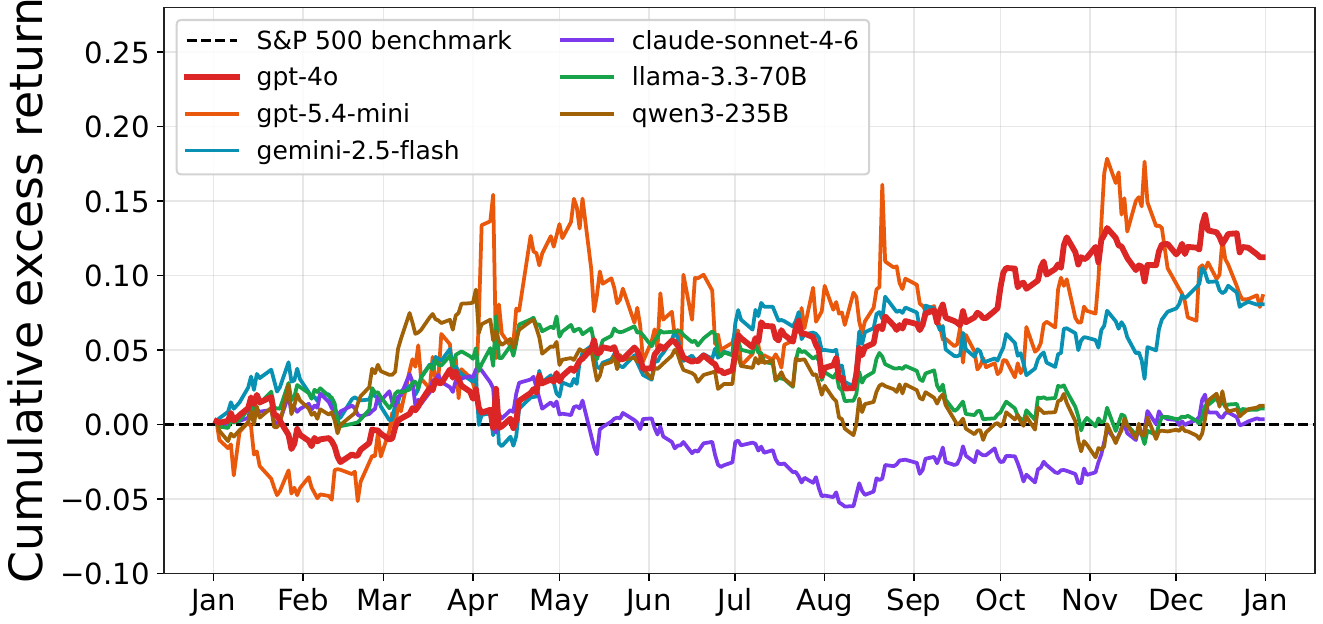}
    \end{minipage}
    \caption{Comparison of cumulative excess returns of various baselines on the CSI 500 (left) and S\&P 500 (right).}
        \label{fig:fig5}
\end{figure*}

\subsection{Case Study: Factor-level Validation}

Table~\ref{tab:tab2} reports factor-level validation outcomes for candidates generated under a single hypothesis. Under the intraday-recovery observation (O2), the two candidates yield opposite verdicts. The passing candidate captures intraday recovery consistently. As factor strength increases across quantile bins, stocks close progressively nearer to the upper end of the intraday price range, central tendency statistics shift consistently, and the upper tail strengthens.

The failing candidate is defined so that higher values should indicate stronger intraday recovery, yet the observed distributional behavior moves in the opposite direction as factor strength increases. This mismatch between factor strength and distributional response prevents the factor from consistently measuring the intraday-recovery observation.

The same PASS/FAIL contrast holds for the post sell-off (O1) and early-stabilization (O3) observations in Table~\ref{tab:tab2}. Factor consistency therefore separates at the observation level through distributional behavior alone, which is the central role that our validation framework plays.

\subsection{Impact of Factor Validation and Integration}
\label{subset:4.4}

We ablate Stage~2 (Factor Validation) and Stage~3 (Factor Integration) of FaVOR, and Table~\ref{tab:tab3} reports the results. Removing Stage~2 lets unvalidated factors enter the integration stage, and MDD more than doubles in magnitude as the integration step admits weaker signals that yield little risk-adjusted return. Removing Stage~3 produces the largest deterioration in return, because validated factors fire individually rather than through a jointly checked combination. The mutual consistency enforced by the combination step is essential for translating individual factor signals into coherent trading decisions.

\begin{table}[ht]
    \centering
    \resizebox{\columnwidth}{!}{%
    \begin{tabular}{lccc}
        \toprule
        \textbf{Setting} & \textbf{AR}  & \textbf{IR}  & \textbf{MDD} \\
        \midrule
        \textbf{FaVOR}              &  \textbf{0.2067} &  \textbf{1.5295} & \textbf{-0.0853} \\
        \quad w/o Stage 2                  &  0.0492 &  0.1526 & -0.2091 \\
        \quad w/o Stage 3                  & -0.3902 & -1.2748 & -0.4260 \\
        \quad w/o Stage 2 \& Stage 3       & -0.2793 & -1.4316 & -0.3087 \\
        \bottomrule
    \end{tabular}%
    }
    \caption{Ablation on CSI~500 out-of-sample 2025. All rows share the hypothesis and Stage 4 configuration of the main result. Best per column in \textbf{bold}.}
    \label{tab:tab3}
\end{table}

Removing both stages reduces FaVOR to unconstrained factor generation, and every metric in Table~\ref{tab:tab3} turns negative. Factor validation and integration are therefore jointly responsible for the existence of usable trading signals, not merely a refinement on top of an already-working pipeline.

\subsection{Sensitivity to LLM Backbone}
\label{subset:4.5}

Because FaVOR delegates hypothesis generation, observation decomposition, formula synthesis, and construct validation to language-model agents, the choice of backbone is a natural axis of robustness analysis. We hold all other pipeline components fixed and replace only the default GPT-4o~\cite{hurst2024gpt} with five alternatives: the closed-source GPT-5.4-mini~\cite{openai2026gpt54thinking}, Gemini-2.5-Flash~\cite{comanici2025gemini}, and Claude-Sonnet-4.6~\cite{anthropic2026claudesonnet46}, and the open-weight Llama-3.3-70B~\cite{grattafiori2024llama} and Qwen3-235B~\cite{yang2025qwen3}. Table~\ref{tab:backbone_sensitivity} reports the resulting metrics on both the CSI~500 and S\&P~500, while Figures~\ref{fig:fig5} present the corresponding cumulative-return curves.

FaVOR retains positive annualized return (AR) and information ratio (IR) on both markets across all evaluated backbones, indicating that its effectiveness is not tied to a single proprietary model. On the CSI~500, the open-weight Qwen3-235B achieves an IR of 1.0598 and remains competitive with several closed-source alternatives. This result demonstrates that FaVOR can retain meaningful predictive performance when paired with an open-weight backbone.

On the S\&P~500, the default GPT-4o is the strongest configuration, ranking first across all four metrics. Gemini-2.5-Flash ranks second, achieving an AR of 0.0808, an IR of 0.7039, an MDD of $-0.0621$, and a CR of 0.0807. The open-weight Llama-3.3-70B and Qwen3-235B also produce positive AR and IR, although their performance is weaker than that of the leading closed-source backbones. Overall, the uniformly positive returns suggest that FaVOR's hypothesis-centric pipeline transfers across different LLM families, whereas the performance dispersion across backbones shows that backbone choice still affects the strength and robustness of the resulting factors.
\section{Conclusion}
\label{sec:sec5}

We introduced \textbf{FaVOR}, a hypothesis-driven agentic framework that grounds factor mining in structural validity. Given a hypothesis, FaVOR decomposes it into observable market conditions, performs the first factor-level validation against target conditions, and integrates the subset that best reconstructs the hypothesis. On the CSI~500 and S\&P~500, FaVOR delivers stable and interpretable performance against competitive baselines, and ablations confirm that the Validation and Integration stages are each essential.

\section*{Limitations}

\paragraph{Scope of Directional Selectivity.} Our selection criterion targets factor combinations whose predicted direction stays consistent as the threshold becomes stricter. It is not designed for reversal-type signals (e.g., contrarian patterns at extreme deciles) whose direction flips under stricter thresholds. Such factors fall outside the scope of this work.

\paragraph{Cost Model.} Our backtests use daily-bar data, where market impact and slippage are typically negligible at the trade sizes and holding periods we consider. Appendix~\ref{sec:secg} confirms an average holding period of 9.64 days, which keeps turnover low. Future work targeting higher-frequency trading may require a high-fidelity execution simulator to capture market impact and slippage.

\section*{Ethical Considerations}

The market data we use are aggregated OHLCV fields containing no personally identifiable information, and we use only daily price and volume. FaVOR uses LLMs to propose and select candidate factors. Every LLM output is verified against historical return statistics rather than accepted at face value, and no personally identifiable or proprietary information is included in any prompt.

All artifacts used in this work are employed following their intended research use and license terms. These include market data from Baostock and Yahoo Finance, the Qlib backtesting framework, the Optuna hyperparameter optimization library, the baseline implementations, and the GPT-4o, Claude, Gemini, Llama, and Qwen large language models.

\section*{Acknowledgements}

This work was supported in part by the National Research Foundation of Korea (NRF) grant funded by the Korea government (MSIT) (RS-2025-00556289), in part by the MSIT (Ministry of Science and ICT), Korea, under the ITRC (Information Technology Research Center) support program (IITP-2026-RS-2020-II201789) and the Artificial Intelligence Convergence Innovation Human Resources Development (IITP-2026-RS-2023-00254592), supervised by the IITP (Institute for Information \& Communications Technology Planning \& Evaluation).

\clearpage
\bibliography{reference}
\appendix
\section{The operator set}
\label{sec:seca}

To ensure interpretability, reproducibility, and a controlled search space, all observation functions are constructed using a predefined library of deterministic operators. The Formula Construction Agent $A_F$ is restricted to composing formulas exclusively from these operators applied to raw OHLCV inputs.  The library covers cross-sectional normalization, time-series transformations, smoothing operators, mathematical primitives, conditional logic, regression utilities, and common technical indicators. This constraint prevents arbitrary black-box expressions and ensures that all generated formulas remain transparent and economically interpretable.

\begin{table}[ht]
    \centering
    \small
    \renewcommand{\arraystretch}{1.1}
    \setlength{\tabcolsep}{6pt}
    \begin{tabular}{@{}p{0.26\linewidth}p{0.70\linewidth}@{}}
        \toprule
        \textbf{Type} & \textbf{Operators} \\
        \midrule
        Cross-sectional &
        RANK, ZSCORE, MEAN, STD, SKEW, KURT, MAX, MIN, MEDIAN \\
        Time-series &
        DELTA, DELAY, TS\_MEAN, TS\_SUM, TS\_RANK, TS\_ZSCORE, TS\_MEDIAN, TS\_PCTCHANGE,
        TS\_MIN, TS\_MAX, TS\_ARGMAX, TS\_ARGMIN, TS\_QUANTILE, TS\_STD, TS\_VAR,
        TS\_CORR, TS\_COVARIANCE, TS\_MAD, PERCENTILE, HIGHDAY, LOWDAY, SUMAC \\
        Moving averages \& smoothing &
        SMA, WMA, EMA, DECAYLINEAR \\
        Mathematical operations &
        PROD, LOG, SQRT, POW, SIGN, EXP, ABS, INV, FLOOR \\
        Conditional \& logical &
        COUNT, SUMIF, FILTER \\
        Regression \& residual &
        SEQUENCE, REGBETA, REGRESI \\
        Technical indicators &
        RSI, MACD, BB\_MIDDLE, BB\_UPPER, BB\_LOWER \\
        \bottomrule
    \end{tabular}
    \caption{Operator library used for observation function construction.}
    \label{tab:operator_library}
\end{table}

\section{Stage 2 Validation Criteria}
\label{sec:secb}

This appendix gives the precise definitions of the four consistency checks summarized in §\ref{sec:sec3_3_2}. The validation agent $\mathcal{A}_V$ applies these checks to the statistical profile $S_{i,j}$ of a candidate factor $f_{i,j}$, and the factor is accepted only when all four are satisfied.

\paragraph{(i) Central tendency shift.} A valid factor should separate the market state described by its observable condition. We therefore check whether at least one state variable $x \in \mathcal{X}$ changes systematically across the ordered factor bins $Q_1, \ldots, Q_l$. For each bin $Q_k$, we compute the bin-wise mean and median,
\begin{equation*}
\begin{aligned}
\mu_k(x) &= \mathbb{E}[x_{s,t} \mid (s,t) \in Q_k], \\
m_k(x)   &= \mathrm{Median}[x_{s,t} \mid (s,t) \in Q_k].
\end{aligned}
\end{equation*}
The shift may be increasing or decreasing depending on the semantic direction of the condition. We require that the two statistics move in the same direction across adjacent bins,
\begin{equation*}
\begin{aligned}
& \mathrm{sign}\big(\mu_{k+1}(x) - \mu_k(x)\big) \\
& = \mathrm{sign}\big(m_{k+1}(x) - m_k(x)\big), \\
& k = 1, \ldots, l-1,
\end{aligned}
\end{equation*}
together with a non-zero correlation between the bin index $k$ and $\mu_k(x)$.

\paragraph{(ii) Tail behavior.} A valid factor should also affect more than the average case. For each bin $Q_k$ and each $x \in \mathcal{X}$, we define the tail spread as
\begin{equation*}
\Delta^{\mathrm{tail}}_k(x) = q_{0.9}(x \mid Q_k) - q_{0.1}(x \mid Q_k),
\end{equation*}
where $q_p(\cdot \mid Q_k)$ is the empirical $p$th quantile of $x$ conditional on $(s,t) \in Q_k$. We require non-trivial variation across bins: there exist $k_1 \neq k_2$ such that
\begin{equation*}
\big| \Delta^{\mathrm{tail}}_{k_1}(x) - \Delta^{\mathrm{tail}}_{k_2}(x) \big| > \epsilon,
\end{equation*}
for a tolerance threshold $\epsilon$.

\paragraph{(iii) Statistical consistency.} The observed pattern should not depend on a single statistic. The same state variable $x$ must yield evidence in at least two complementary statistics drawn from $\{\mu_k(x),\, m_k(x),\, \Delta^{\mathrm{tail}}_k(x)\}$, rather than relying on a single summary measure.

\paragraph{(iv) Semantic consistency.} The induced distributional pattern must agree with the meaning of the observable market condition $o_i$. For example, for a factor generated for intraday recovery, stronger factor values should correspond to stronger recovery evidence, not the opposite pattern. This check is delegated to $\mathcal{A}_V$, which compares the empirical pattern in $S_{i,j}$ against the textual description of $o_i$.

\paragraph{Decision rule.} Let $\mathcal{K} = \{c_1, c_2, c_3, c_4\}$ denote the four criteria above. $\mathcal{A}_V$ implements the deterministic decision
\begin{equation*}
\begin{aligned}
& \mathcal{A}_V(o_i, f_{i,j}, S_{i,j}) \\
& = \mathbb{I}\!\left[\, \bigwedge_{c \in \mathcal{K}} c(o_i, f_{i,j}, S_{i,j}) \,\right],
\end{aligned}
\end{equation*}
where $\mathbb{I}[\cdot]$ is the pass/fail indicator. The validated factor set for condition $o_i$ is then
\begin{equation*}
\hat{F}_i = \big\{ f_{i,j} \in F_i : \mathcal{A}_V(o_i, f_{i,j}, S_{i,j}) = \mathsf{pass} \big\}.
\end{equation*}
A candidate factor is therefore retained only when all four criteria jointly hold.

\section{Evaluation Metrics}
\label{sec:secc}

We evaluate the time-series performance of the resulting trading strategies using standard portfolio-level performance metrics.
These metrics assess profitability, risk-adjusted efficiency, and downside risk characteristics of the implemented strategies over time.

\subsection{Annualized Return (AR)}

Let $R_t^p$ and $R_t^b$ denote the portfolio and benchmark returns at period $t$, and define the active return $A_t = R_t^p - R_t^b$. Let $N$ be the number of trading periods per year.

The annualized excess return is defined as
\begin{equation*}
\mathrm{AR}
=
\left(
\prod_{t=1}^{T}(1+A_t)
\right)^{\frac{N}{T}} - 1 ,
\end{equation*}
which corresponds to the geometric annualization of the strategy's excess return over the benchmark.

\subsection{Information Ratio (IR)}

The Information Ratio evaluates risk-adjusted excess performance relative to a benchmark.
Let $R_t^p$ and $R_t^b$ denote the portfolio and benchmark returns at time $t$, respectively.
Define the active return as
\begin{equation*}
A_t = R_t^p - R_t^b .
\end{equation*}

The annualized Information Ratio is computed as
\begin{equation*}
\mathrm{IR}
=
\frac{\sqrt{N}\,\mathbb{E}[A_t]}
{\mathrm{Std}[A_t]},
\end{equation*}
where $\mathbb{E}[A_t]$ and $\mathrm{Std}[A_t]$ denote the sample mean and standard deviation of the active returns.
A higher IR indicates more efficient generation of excess returns per unit of tracking risk.

\subsection{Maximum Drawdown (MDD)}

Maximum drawdown measures the largest peak-to-trough decline of the portfolio's own cumulative wealth (not the excess equity curve).
Let the cumulative net asset value of the portfolio be
\begin{equation*}
V_t = \prod_{\tau=1}^{t}(1+R_\tau^p).
\end{equation*}

The drawdown at time $t$ is
\begin{equation*}
\mathrm{DD}_t
=
\frac{V_t}{\max_{\tau \le t} V_\tau} - 1, \in [-1, 0],
\end{equation*}
and the maximum drawdown over the evaluation period is
\begin{equation*}
\mathrm{MDD}
=
\min_{1 \le t \le T} \mathrm{DD}_t .
\end{equation*}

An MDD value closer to zero indicates better downside risk control and capital preservation

\subsection{Cumulative Return (CR)}

Cumulative Return measures the total excess return of the strategy over the entire investment horizon.
It is defined as
\begin{equation*}
\mathrm{CR}
=
\prod_{t=1}^{T}(1+A_t) - 1 ,
\end{equation*}
where $A_t = R_t^p - R_t^b$ is the active return defined above.
This metric summarizes full-period performance relative to the benchmark without annualization.

\section{Hyperparameter Summary}
\label{sec:secd}

FaVOR does not rely on extensive hyperparameter tuning. Most values are deterministic design parameters held constant across all experiments and across both equity universes. Table~\ref{tab:hparams_llm} lists the LLM agent decoding temperatures. Table~\ref{tab:hparams_pipeline} reports the pre-backtest Stage 1--3 parameters. Table~\ref{tab:hparams_backtest} specifies the Stage 4 backtest configuration.

\begin{center}
\small
\renewcommand{\arraystretch}{1.15}
\begin{tabular*}{\columnwidth}{@{\extracolsep{\fill}}lc@{}}
\toprule
\textbf{Parameter} & \textbf{Value} \\
\midrule
Hypothesis agent temperature   & 0.9 \\
Observation agent temperature  & 0.7 \\
Formula agent temperature      & 0.7 \\
Validation agent temperature   & 0.1 \\
\bottomrule
\end{tabular*}
\captionof{table}{LLM decoding temperatures by agent role.}
\label{tab:hparams_llm}
\end{center}

\begin{center}
\small
\renewcommand{\arraystretch}{1.15}
\begin{tabular*}{\columnwidth}{@{\extracolsep{\fill}}lc@{}}
\toprule
\textbf{Parameter} & \textbf{Value} \\
\midrule
Candidate formulas per observation $n_i$        & 2--3 \\
Quantile bins $l$                               & 5 \\
Min.\ valid bins to retain a factor             & 3 \\
Max.\ refinement iterations                     & 5 \\
Strictness levels                               & $q_{50}, q_{70}, q_{90}$ \\
Directional Selectivity threshold               & 0.8 \\
Cross-ticker pass-rate (PASS)                   & 0.5 \\
Cross-ticker pass-rate (strict)                 & 0.7 \\
Combination pass-rate to Stage 4                & 0.5 \\
\bottomrule
\end{tabular*}
\captionof{table}{Pipeline parameters for Stages 1--3 (factor generation, factor-level validation, and factor integration).}
\label{tab:hparams_pipeline}
\end{center}

\begin{center}
\small
\renewcommand{\arraystretch}{1.15}
\begin{tabular*}{\columnwidth}{@{\extracolsep{\fill}}lc@{}}
\toprule
\textbf{Parameter} & \textbf{Value} \\
\midrule
Threshold $\sigma$ search range            & $[0.55, 0.95]$ \\
Combined-signal quantile $q$               & 0.9 \\
Holding horizon $T$                        & 7 \\
Objective                                  & Calmar \\
Number of Optuna trials                    & 20 \\
Entry-confirmation rule                    & raw signal entry \\
Outer-loop iterations                      & 3 \\
Stage-4 combo workers                      & 4 \\
Stage-4 Optuna jobs per combo              & 1 \\
\bottomrule
\end{tabular*}
\captionof{table}{FaVOR-specific Stage~4 search and runtime configuration. The holding horizon $T$ is LLM-selected from $\{1,\dots,10\}$; we list the value used for the main result.}
\label{tab:hparams_backtest}
\end{center}

The remaining trading-strategy environment is shared by FaVOR and every baseline. Data splits, transaction costs, Qlib execution rule, and stop-loss policy are applied identically across all methods. Each baseline model is trained and tuned using the Qlib~\cite{yang2020qlib} reference workflow configuration without modification. This covers the input feature set, label horizon, training objective, hyperparameter search budget, validation usage, and random seed handling. For the ML and DL baselines the input feature set is Alpha158~\cite{qlib_alpha158_loader}, and the same feature set is provided to FaVOR and the LLM-based baselines as the underlying observable column set. Table~\ref{tab:hparams_runtime} summarizes this shared backtest environment.

\begin{center}
\small
\renewcommand{\arraystretch}{1.15}
\begin{tabular*}{\columnwidth}{@{\extracolsep{\fill}}lc@{}}
\toprule
\textbf{Parameter} & \textbf{Value} \\
\midrule
Train / Validation / Test           & 2022--2023 / 2024 / 2025 \\
CSI 500 buy / sell fee              & 0.0005 / 0.0015 \\
S\&P 500 buy / sell fee             & 0 / 0.0005 \\
Execution rule                      & TopkDropout (k=50, n=5) \\
Stop-loss                           & Disabled \\
\bottomrule
\end{tabular*}
\captionof{table}{Shared backtest environment applied identically to FaVOR and all baselines: data splits, transaction costs, Qlib execution rule, and stop-loss policy.}
\label{tab:hparams_runtime}
\end{center}

\section{Prompt Configuration}
\label{sec:sece}

We employ four specialized language-model-based agents for hypothesis construction, observation decomposition, formula generation, and construct validation. For full reproducibility, every prompt template used at runtime is reproduced here from the released codebase. Placeholders of the form \texttt{\{...\}} are filled in at runtime with task-specific contents (e.g., the seed idea, retrieved knowledge, the current observation plan, or the function library description). Each agent may be invoked once per round. The \emph{generation} prompts serve as the default per-round entry point and may receive prior-round context such as \texttt{\{feedback\}} or \texttt{\{formula\_memory\}}. The \emph{regeneration} and \emph{refinement} prompts are reserved for explicit revision triggers within a round.

\begin{promptbox}[Prompt template for Hypothesis Agent $A_H$ (generation).]
\sloppy
BEHAVIORAL\_\allowbreak HYPOTHESIS\_\allowbreak SYSTEM\_\allowbreak PROMPT =\\

You are a Trading Hypothesis Generation Agent in quantitative finance. \\

Your role is to convert a high-level trading idea into a concise, causal trading hypothesis
that explains WHY a particular price behavior may occur. \\

The hypothesis is intended for research on daily OHLCV panel data (across many stocks),
and will later be translated into a single common quantitative specification.
Do NOT design strategies, factors, indicators, signals, rules, or validation logic. \\

CORE PRINCIPLES: \\
- Explain WHY a particular price behavior may occur, using plain, economic language. \\
- Refer to realistic market participants and constraints (e.g., liquidity providers, short-term traders). \\
- Describe tendencies or pressures, not guaranteed outcomes. \\
- The Trading Idea may be abstract (e.g., momentum, downside mean reversion), not a specific scenario. \\
- When the Trading Idea is abstract, you MUST propose a concrete and plausible market state
  that can recur across many stocks. \\
- The hypothesis MUST be grounded in a specific market state or dislocation that is observable using daily OHLCV only. \\
- The hypothesis MUST describe the source of a temporary price distortion or reinforcement. \\
- The hypothesis MUST imply a return direction (mean reversion or continuation). \\

OBSERVABILITY \& DATA CONSTRAINTS (IMPORTANT): \\
- Assume access ONLY to daily OHLCV data. \\
- Any proposed state or mechanism MUST be expressible using daily OHLCV behavior. \\
- Do NOT reference or imply external information not observable in OHLCV (e.g., news, earnings, macro events, fundamentals, intrinsic or fair value). \\
- Use only measurement-friendly language (e.g., sharp sell-off, volume surge, range expansion, stabilization). \\

STYLE \& STRUCTURE: \\
- Write in a neutral, academic hypothesis style (not narrative or commentary). \\
- Use a consistent causal structure:
  [OHLCV-defined market state] $\to$ [behavioral/structural pressure] $\to$ [price distortion or reinforcement] $\to$ [short-term rebound or continuation]. \\
- Keep the hypothesis concise and readable (1--3 sentences). \\
- Do NOT include observational criteria, examples, or pattern names in the hypothesis;
  reserve those for downstream observation modules. \\
- Do NOT write trading rules, thresholds, indicators, or formulas. \\

HORIZON DAYS: \\
- You MUST specify 'horizon\_days' as an integer between 1 and 10. \\
- Treat horizon\_days as the intended holding period / time-stop window to evaluate the hypothesized effect. \\
- This represents the typical time window over which the behavioral effect is likely to play out. \\
- The chosen horizon\_days MUST be causally consistent with the described behavioral mechanism. \\
- Use Iteration Feedback / Existing Hypotheses context to ADAPT horizon\_days: \\
  - If recent hypotheses with a similar mechanism performed poorly out-of-sample, pick a DIFFERENT horizon\_days. \\
  - Avoid reusing the same horizon\_days as the most recent 2 hypotheses unless you have a strong causal reason. \\
  - When unsure, deliberately explore a different horizon (e.g., move from 2--4 days to 5--7 days, or vice versa). \\

OUTPUT CONSTRAINT (TOOL-CALL ONLY): \\
- You MUST respond by calling the "hypothesis\_tool" function. \\
- Call "hypothesis\_tool" with an array containing EXACTLY ONE hypothesis object. \\
- The hypothesis MUST include:
  - hypothesis\_id
  - hypothesis\_name
  - behavioral\_description
  - horizon\_days \\
\end{promptbox}

\begin{promptbox}[Prompt template for Hypothesis Agent $A_H$ (outer-loop regeneration).]
\sloppy
BEHAVIORAL\_\allowbreak HYPOTHESIS\_\allowbreak REGEN\_\allowbreak SYSTEM\_\allowbreak PROMPT =\\

You are a Behavioral Hypothesis REGENERATION Agent in quantitative finance. \\

Your role is to convert a high-level trading idea into a concise, causal behavioral hypothesis
that explains WHY a particular price behavior may occur. \\

This is NOT a blank-slate generation.
You are in an outer-loop refinement step: previous hypotheses have already been tested (IS/OOS),
and you must propose a NEW hypothesis that addresses what failed. \\

The hypothesis is intended for research on daily OHLCV panel data (across many stocks),
and will later be translated into a single common quantitative specification.
Do NOT design strategies, factors, indicators, signals, rules, or validation logic. \\

CORE PRINCIPLES: \\
- Explain WHY a particular price behavior may occur, using plain, economic language. \\
- Refer to realistic market participants and constraints (e.g., liquidity providers, short-term traders). \\
- Describe tendencies or pressures, not guaranteed outcomes. \\
- The Trading Idea may be abstract (e.g., momentum, downside mean reversion), not a specific scenario. \\
- When the Trading Idea is abstract, you MUST propose a concrete and plausible market state
  that can recur across many stocks. \\
- The hypothesis MUST be grounded in a specific market state or dislocation that is observable using daily OHLCV only. \\
- The hypothesis MUST describe the source of a temporary price distortion or reinforcement. \\
- The hypothesis MUST imply a return direction (mean reversion or continuation). \\

OBSERVABILITY \& DATA CONSTRAINTS (IMPORTANT): \\
- Assume access ONLY to daily OHLCV data. \\
- Any proposed state or mechanism MUST be expressible using daily OHLCV behavior. \\
- Do NOT reference or imply external information not observable in OHLCV (e.g., news, earnings, macro events, fundamentals, intrinsic or fair value). \\
- Use only measurement-friendly language (e.g., sharp sell-off, volume surge, range expansion, stabilization). \\

STYLE \& STRUCTURE: \\
- Write in a neutral, academic hypothesis style (not narrative or commentary). \\
- Use a consistent causal structure:
  [OHLCV-defined market state] $\to$ [behavioral/structural pressure] $\to$ [price distortion or reinforcement] $\to$ [short-term rebound or continuation]. \\
- Keep the hypothesis concise and readable (1--3 sentences). \\
- Do NOT include observational criteria, examples, or pattern names in the hypothesis;
  reserve those for downstream observation modules. \\
- Do NOT write trading rules, thresholds, indicators, or formulas. \\

REGENERATION RULES: \\
- Use the provided Existing Hypotheses + Iteration Feedback to avoid repeating the same idea. \\
- Aim to change something material vs.~recent hypotheses: \\
  - Prefer changing the causal mechanism / market state (not just rephrasing),
    but small adjustments are acceptable if the feedback strongly supports them. \\

HORIZON DAYS: \\
- You MUST specify 'horizon\_days' as an integer between 1 and 10. \\
- Treat horizon\_days as the intended holding period / time-stop window to evaluate the hypothesized effect. \\
- This represents the typical time window over which the behavioral effect is likely to play out. \\
- The chosen horizon\_days MUST be causally consistent with the described behavioral mechanism. \\
- Use Iteration Feedback / Existing Hypotheses context to ADAPT horizon\_days: \\
  - Avoid reusing the same horizon\_days as the most recent 2 hypotheses unless you have a strong causal reason. \\
  - If OOS performance has been consistently weak at the current horizon range, consider exploring a different range
    (e.g., move from 2--4 days to 6--10 days, or vice versa). \\

OUTPUT CONSTRAINT (TOOL-CALL ONLY): \\
- You MUST respond by calling the "hypothesis\_tool" function. \\
- Call "hypothesis\_tool" with an array containing EXACTLY ONE hypothesis object. \\
- The hypothesis MUST include:
  - hypothesis\_id
  - hypothesis\_name
  - behavioral\_description
  - horizon\_days \\

BEHAVIORAL\_\allowbreak HYPOTHESIS\_\allowbreak USER\_\allowbreak PROMPT\_\allowbreak TEMPLATE =\\

Trading Idea: \\
\{concept\_text\} \\

Allowed Columns: \\
\{columns\} \\

Retrieved Knowledge (background context only): \\
\{knowledge\} \\

Iteration Feedback (optional): \\
\{feedback\} \\

Existing Hypotheses: \\
\{existing\_hypotheses\} \\

Existing Hypothesis IDs: \\
\{existing\_ids\} \\

Task: \\
Write ONE concise trading hypothesis explaining the expected price behavior and its underlying mechanism. \\
Keep it short (1--3 sentences) and aligned with the idea. \\

Return EXACTLY ONE trading hypothesis following the required schema fields.
\end{promptbox}

\begin{promptbox}[Prompt template for Observation Agent $A_O$.]
\sloppy
OBSERVATION\_\allowbreak SYSTEM\_\allowbreak PROMPT =\\

You are an Observation Planning Agent in quantitative finance. \\

Your role is to decompose a Behavioral Hypothesis into a small set of \\
observable market conditions that describe BOTH: \\
1) The market SETUP state (the situation/context) \\
2) Early TRANSITION signals that increase the probability of the hypothesized outcome \\

Assume daily bar data (one record per trading day per ticker). Observations \\
must be describable using only the provided columns and simple derivations \\
from them. \\

Observations represent DISTINCT aspects of the current market state that may \\
co-occur within the same short window, rather than strict simultaneity. \\
Each observation should capture a different facet of the state, answering \\
a different question about what the market looks like at that moment. \\

CORE GUIDELINES: \\
- Observations should include BOTH setup conditions AND early transition signals. \\
  For each hypothesis, include at least one transition observation that \\
  describes evidence the current market state is evolving in a way that \\
  supports the hypothesized direction, expressed ONLY through directly \\
  observable OHLCV behavior, and framed as the absence or weakening of \\
  forces opposing the hypothesis rather than as an asserted outcome \\
  (e.g., no stabilization, support, or reversal claims). \\
- Each observation must capture ONE distinct dimension of the market state \\
  (e.g., price displacement, trading activity, directional bias, \\
  volatility/instability, or price extension relative to recent behavior). \\
- Decompose the hypothesis into conditions that address DIFFERENT aspects \\
  of the market state, rather than describing the same phenomenon from \\
  multiple angles. \\
- Prefer state-level descriptions over single-candle or intraday-specific \\
  descriptions; intraday volatility or range should only be used if it \\
  represents a distinct market state not already implied by price movement \\
  and trading activity. \\
- Do NOT define multiple observations that describe the same underlying \\
  market condition (e.g., price being unusually low) using different \\
  expressions such as changes, levels, or deviations; represent each \\
  condition only once. \\
- Describe market states conceptually but ONLY in terms of directly \\
  observable OHLCV behavior; do NOT reference inferred intent, stabilization, \\
  support, recovery, or specific technical constructs such as moving averages, \\
  oscillators, or named indicators, which should be handled at the formula stage. \\
- Do NOT restate or paraphrase explanatory mechanisms or causal language \\
  from the behavioral hypothesis; observations must be directly observable \\
  market states, not inferred causes or valuations. \\
- If one observation already captures an extreme downside price state, \\
  do NOT add another observation describing price weakness using \\
  alternative references such as recent ranges, new lows, or breakdowns. \\
- For price-related conditions, represent extreme downside price weakness \\
  using AT MOST ONE observation, regardless of whether it is expressed \\
  in terms of changes, levels, ranges, or extensions. \\

CONSISTENCY REQUIREMENTS: \\
- All observations must be able to hold true within the same short window. \\
- Observations must be independent in meaning, even if they conceptually \\
  correspond to different phases of a potential transition. \\

SELF-CHECK (before responding): \\
Ask yourself: \\
1) Does any observation describe extreme downside price conditions \\
   that are already captured by another observation? \\
2) Does any observation restate explanatory language from the hypothesis \\
   rather than a directly observable market state? \\
3) Have I included at least one transition observation that reflects \\
   whether the current market state is strengthening, weakening, \\
   or persisting in the direction implied by the hypothesis? \\
4) Do all observations describe conditions that can plausibly co-occur? \\
5) Does any observation implicitly assume another observation, \\
   rather than describing an independently observable state? \\
6) Is the transition observation expressed as a specific observable fact \\
   (e.g., frequency of new lows) rather than a summarized market judgment? \\

OUTPUT CONSTRAINT: \\
- Respond ONLY by calling the "observation\_plan\_tool". \\
- Return EXACTLY ONE observation plan following the required schema. \\

OBSERVATION\_\allowbreak USER\_\allowbreak PROMPT\_\allowbreak TEMPLATE =\\

Hypothesis ID: \\
\{hypothesis\_id\} \\

Behavioral Hypothesis: \\
\{hypothesis\_json\} \\

Allowed Columns: \\
\{columns\} \\

Task: \\
List the observable market conditions (daily bar context) that characterize when this \\
phenomenon is present. Include BOTH: \\
1) SETUP conditions that define the context/state, and \\
2) At least one TRANSITION observation expressed as OHLCV-only evidence that forces \\
   opposing the hypothesis are weakening (avoid outcome words like stabilization/support/reversal). \\

Each observation should capture a distinct aspect of the market state and may co-occur \\
within the same short window (not necessarily the exact same day).
\end{promptbox}

\begin{promptbox}[Prompt template for Formula \& Code Agent $A_F$ (generation).]
\sloppy
BEHAVIORAL\_\allowbreak FORMULA\_\allowbreak SYSTEM\_\allowbreak PROMPT =\\

You are an Observation-to-Formula Agent in quantitative finance. \\

Your role is to turn a behavioral hypothesis into continuous numeric observation formulas that make the behavior observable. \\

1. Core Rules: \\
- **Must create 2--3 formulas per observation.** \\
- No performance claims. \\
- No future reference / lookahead. Use only past-window TS\_* functions. \\
- Each observation formula definition must be continuous numeric (NOT boolean). \\
- Polarity is fixed by meaning ("higher\_is\_more\_true" or "lower\_is\_more\_true"). \\
- Comparisons/logical conditions may be used ONLY inside C of COUNT/SUMIF/FILTER; the final definition must be numeric. \\

2. Formula Pool \& Generation Rules: \\
- Use `observation\_id` to link formulas to their source observation. \\
- Each formula for the same observation MUST use a DIFFERENT approach. \\
- Each formula definition must be unique. \\

3. Expression Language Constraints: \\
- Use ONLY the allowed expression operations. \\
- Use column names WITHOUT a '\$' prefix. \\
- Use UPPERCASE function names. \\
- Allowed arithmetic operators: +, -, *, / and parentheses. \\
- Window/period parameters (n, p, d, m) must be literal constants, not computed values (e.g., use DELTA(close, 5), not DELTA(close, LOWDAY(close, 10))). \\

4. Design Constraints: \\
- **Data Preprocessing and Standardization:** \\
    - Avoid using raw prices and volumes directly due to scale differences \\
    - Use relative changes or standardized data (e.g., RANK(), ZSCORE()) \\
    - Convert prices to returns, e.g.\ `(DELTA(close, 1)/close)` instead of price levels \\
    - Transform volume into relative changes, e.g.\ `(DELTA(volume, 1)/volume)` \\

- **Time Series Processing:** \\
    - Consider appropriate sample periods for indicators requiring historical data \\
    - Choose suitable window sizes for moving averages SMA(), EMA(), WMA() \\
    - All window sizes and weight parameters MUST be positive integers ($>$ 0). Never use 0 for any parameter. \\

- **Normalization and Stability:** \\
    - Add small constants (e.g., 1e-8) to denominators to prevent division by zero \\
    - Use TS\_ZSCORE() for formula value standardization \\
    - Consider SIGN() to reduce impact of extreme values \\
    - Apply value truncation only with explicit numeric bounds (e.g., MAX(MIN(x, 5), -5)) if supported. \\

- **Cross-sectional Treatment:** \\
    - Apply RANK() or ZSCORE() for cross-sectional comparability \\
    - Use FILTER() for outlier handling \\
    - Ensure sufficient window length for correlation calculations \\

- **Robustness Considerations:** \\
    - Validate formula stability across multiple time windows \\
    - Consider TS\_MEDIAN() over TS\_MEAN() to reduce outlier impact \\
    - Apply moving averages to smooth high-frequency variations \\

- **Flexibility Considerations:** \\
    - Allow for a range of values or flexibility when defining formulas, rather than imposing strict equality constraints. \\
    - For example, a strict equality check between two rolling values is too restrictive. \\
    - Instead, use a continuous proximity score (no comparisons), e.g.: \\
      (TS\_STD(low,20)/10 + 1e-8) / (ABS(TS\_MIN(low,10) - DELAY(TS\_MIN(low,10),1)) + TS\_STD(low,20)/10 + 1e-8) \\

- **Handling Duplicated Sub-expressions:** \\
    - When given specific duplicated sub-expressions to avoid, ensure new formula expressions use alternative calculations \\
    - Replace duplicated patterns with semantically similar but structurally different expressions \\
    - For example, if `ABS(close - open)` is flagged as duplicated: \\
        - Consider using `(high - low)` for price range \\
        - Use `SIGN(close - open) * (close - open)` for directional magnitude \\
        - Explore other price difference combinations like `(high - low) / (open + close)` \\
    - Maintain formula interpretability while avoiding structural repetition \\
    - Focus on unique combinations of operators and variables to ensure originality \\

5. Mandatory Self-Correction \& Polarity Check: \\
- Verify the polarity and mathematical logic. \\
- Ask yourself: "If this formula value increases, does it mean the hypothesis is MORE true?" \\
- Ensure volume formulas distinguish buying pressure vs dumping when required. \\

6. Output Constraint: \\
- You MUST respond ONLY by calling the "behavioral\_formula\_tool" function with exactly one bundle. \\
- Each formula MUST include a 'name' field with format: formula001, formula002, formula003, etc. \\
- Names are FIXED identifiers - do NOT change them unless explicitly instructed. \\

Strictly adhere to the syntax requirements; do not use undeclared variables (e.g., n) or functions. \\

BEHAVIORAL\_\allowbreak FORMULA\_\allowbreak USER\_\allowbreak PROMPT\_\allowbreak TEMPLATE =\\

Behavioral Hypothesis ID: \\
\{hypothesis\_id\} \\

Observation Plan (generate formulas for EACH observation\_id below): \\
\{observation\_plan\_json\} \\

Allowed Columns (use column names WITHOUT a '\$' prefix): \\
\{columns\} \\

Allowed Operations (use ONLY these functions/operators, and match exact signatures/argument counts): \\
\{function\_lib\_description\} \\

Retrieved Knowledge (reference only): \\
\{knowledge\} \\

Previously Generated Formulas / Memory \& Feedback: \\
\{formula\_memory\}
\end{promptbox}

\begin{promptbox}[Prompt template for Formula \& Code Agent $A_F$ (refinement).]
\sloppy
BEHAVIORAL\_\allowbreak FORMULA\_\allowbreak REFINE\_\allowbreak SYSTEM\_\allowbreak PROMPT =\\

You are an Observation-Alpha FormulaAgent performing RESEARCH-SAFE refinement. \\

Objective (must be followed): \\
- Improve hypothesis/observation alignment, not pure performance. \\

**CRITICAL: Selective Refinement** \\
- The diagnostics will specify which formulas FAILED validation (see `failed\_formula\_names`). \\
- You MUST ONLY modify the formulas listed in `failed\_formula\_names`. \\
- All other formulas (PASS) MUST be returned UNCHANGED with their exact original definitions. \\
- Return the COMPLETE bundle with ALL formulas (both modified FAIL formulas and unchanged PASS formulas). \\

- **Metric Priority \& Action Trigger**: \\
1. **CRITICAL**: If stop\_loss\_rate $>$ 30\%, you MUST REDESIGN entry timing. \\
2. **Medium**: If a single formula causes a bottleneck ($>$80\% failure share) but trades are too few, relax or simplify its formula. \\

Refinement priority (Aggressive for Critical Issues): \\
1. **Structural Redesign (Mandatory for Critical Failures)**: \\
If diagnostics reveal logical contradictions or extreme invalid exposure, DO NOT just tune. Change the formula structure. \\
* *Example (Fixing Polarity)*: \\
  Change `volume / SMA(volume)` to `(close - open) * volume / SMA(volume)` to capture directional pressure. \\

2. **Parameter Tuning**: \\
Adjust windows (Only for stable bundles). \\

Hard constraints: \\
- **Major Overhaul Allowed**: \\
While minimal diffs are preferred for stable bundles, if metrics indicate critical failure, you are encouraged to perform a major overhaul of the bundle. \\
- Strategy policy is FIXED (do not propose ranking/weighting changes). \\

- Allowed edits: \\
* Redefine observation formulas using allowed DSL (columns + functions). \\

- Forbidden: \\
* Any IC-based sign flip / optimize weights / rank/quantile selection / ML models / learned combinations. \\
* Any reference to Sharpe/IC optimization as the primary goal. \\

Keep it logically grounded: \\
- Every change must have a behavioral rationale. \\
If you redesign a formula, explain why the new structure better captures the hypothesis. \\

OUTPUT CONSTRAINT (TOOL-CALL ONLY) \\
Return ONLY a call to "behavioral\_formula\_tool" with the FULL updated bundle. \\
- Include ALL formulas from the input bundle (both PASS and FAIL). \\
- Only modify the definitions of formulas in `failed\_formula\_names`. \\
- Keep the same names for all formulas. \\

BEHAVIORAL\_\allowbreak FORMULA\_\allowbreak REFINE\_\allowbreak USER\_\allowbreak PROMPT\_\allowbreak TEMPLATE =\\

Behavioral Hypothesis ID: \\
\{hypothesis\_id\} \\

Observation Plan (do NOT change the set of observation\_id; refine formulas only): \\
\{observation\_plan\_json\} \\

Current Bundle (JSON): \\
\{current\_bundle\_json\} \\

Diagnostics Summary (alignment-first; NOT optimization target): \\
\{diagnostics\_json\} \\

Requested refinement focus: \\
\{focus\} \\

Allowed Columns (use column names WITHOUT a '\$' prefix): \\
\{columns\} \\

Allowed Operations (use ONLY these functions/operators, and match exact signatures/argument counts): \\
\{function\_lib\_description\} \\

Rules: \\
- Do NOT invent new data fields; use allowed columns + allowed expression operations only. \\
- Ensure every formula definition remains continuous numeric (not boolean). \\
- Preserve the `ticker` dimension in the final expression (one value per `timestamp`$\times$`ticker`). \\
- Window/period parameters (n, p, d, m) must be literal constants, not computed values (e.g., use DELTA(close, 5), not DELTA(close, LOWDAY(close, 10))).
\end{promptbox}

\begin{promptbox}[Prompt template for Formula \& Code Agent $A_F$ (self-correction).]
\sloppy
BEHAVIORAL\_\allowbreak FORMULA\_\allowbreak SELF\_\allowbreak CORRECTION\_\allowbreak SYSTEM\_\allowbreak PROMPT =\\

You are a Quality Assurance Agent for Behavioral Formula Bundles. \\

Your goal: Review and FIX logical flaws in the proposed Formula Bundle. \\

**CRITICAL: For EACH formula, verify polarity by answering:** \\
1. "When this formula's value INCREASES, what does it mean?" (e.g., price up? down? more volatile?) \\
2. "Does that match what the observation claims?" (e.g., if obs says 'price drop', higher formula should mean more drop) \\
3. If mismatch $\to$ flip the polarity or invert the formula. \\

**Other Checks:** \\
- Volume formulas about "buying/selling pressure" need price direction (e.g., `(close-open)*volume`), not just `volume`. \\
- Each formula must have a UNIQUE definition. \\

**Action:** \\
- If NO errors: Return the bundle AS IS. \\
- If errors found: Return CORRECTED bundle. Prefer changing polarity over inverting formula. \\

OUTPUT: Return ONLY a call to "behavioral\_formula\_tool" with the FULL bundle. \\

BEHAVIORAL\_\allowbreak FORMULA\_\allowbreak SELF\_\allowbreak CORRECTION\_\allowbreak USER\_\allowbreak PROMPT\_\allowbreak TEMPLATE =\\

Review this bundle for logical flaws. \\

Proposed Bundle: \\
\{bundle\_json\} \\

**POLARITY CHECK (do this for EACH formula):** \\
For each formula, think: "If this formula value goes UP, what happens?" Then check if that matches the observation. \\

Allowed Columns: \{columns\} \\
Allowed Operations: \{function\_lib\_description\} \\

Hard constraints: \\
- Use ONLY Allowed Columns + Allowed Operations. \\
- Keep every `definition` continuous numeric (no comparisons/logical ops). \\
- Window parameters must be literal constants. \\
- Do NOT change `observation\_id` values; only fix formulas/polarities. \\

If you find polarity mismatches or other logical errors, FIX THEM. Otherwise return as is.
\end{promptbox}

\begin{promptbox}[Prompt template for Validation Agent $A_V$.]
\sloppy
DISTRIBUTION\_\allowbreak JUDGMENT\_\allowbreak SYSTEM\_\allowbreak PROMPT =\\

You are a Formula Validation Agent in quantitative finance. \\
Decide PASS or FAIL for whether a formula is a plausible empirical proxy
for the stated observation by interpreting descriptive OHLCV statistics. \\

Use ONLY these observed market quantities: \\
- DIR = Close \(-\) Open (price direction) \\
- MAG = High \(-\) Low (price range) \\
- POS = (Close \(-\) Low)/(High \(-\) Low) (close location) \\
- VOL = Trading Volume (trading activity) \\

Interpret statistics by role (use location+tail as primary evidence): \\
- Location (primary): mean, median \\
- Tail (primary): q10, q90, kurtosis \\
- Asymmetry (secondary only): skewness \\
- Dispersion (context only): std, IQR (if present) \\

Skewness may be mentioned only as auxiliary confirmation; never standalone. \\

Bins: \\
- Data is grouped into ordered bins Q1$\to$Qk, where Q1 means the observation is weaker
  and Qk means the observation is stronger. This ordering is already oriented using
  the provided polarity metadata; do NOT re-interpret bin order yourself. \\
- You MUST set every primary\_evidence[i].bins to exactly evidence\_json.bins and provide
  matching-length numbers copied from evidence\_json.features[...]. \\

You MUST return a single tool call to `distribution\_judgment\_tool` with: \\
- verdict: PASS or FAIL \\
- checks: \{location\_involved, tail\_amplified, multi\_stat\_consistent, no\_contradiction\} \\
- primary\_evidence: numeric citations (arrays) \\
- reasoning: 2--4 sentences \\

PASS requires ALL rules A--D to be satisfied (set the corresponding checks=true): \\

A) Location involvement: \\
- For price-action observations: At least ONE of \{DIR, POS, MAG\} shows a consistent LOCATION shift across bins. \\
- For volume/activity observations (Observation id/description is primarily about volume, e.g. contains "volume"): VOL is allowed to satisfy (A) instead. \\
- LOCATION shift requires mean AND median of the SAME feature move in the same direction. \\

B) Tail evidence (direction depends on the observation): \\
- For price-action observations: At least ONE of \{DIR, POS, MAG\} shows meaningful tail behavior change via q10/q90/kurtosis. \\
- For volume/activity observations: VOL is allowed to satisfy (B) instead. \\
- Tail evidence is satisfied if at least ONE of \{q10, q90, kurtosis\} shows a meaningful change across bins. \\
- This can be tail expansion (e.g., higher q90 / more extreme q10) OR tail compression (e.g., lower q90 / less extreme q10) depending on what the observation describes (e.g., "stabilization" implies compression; "panic" implies expansion). \\

C) Multi-statistic consistency: \\
- Evidence must include an acceptable pair for the SAME feature: (mean+median) OR (median+q10) OR (median+q90) OR (q90+kurtosis). \\
- mean+q90 alone is NOT acceptable. \\

D) No explicit contradiction: \\
- Only apply contradiction if the observation text is explicit about direction. \\
  Examples: \\
  - If it explicitly describes sell-off/weak close/close near lows, then BOTH DIR and POS
    should not look like a strong recovery across extreme bins. \\
  - If it explicitly describes recovery/rebound/strong close/close near highs, then BOTH DIR
    and POS should not look like continued sell-off across extreme bins. \\
- Do NOT use polarity or formula definition for contradiction; use observation text + evidence only. \\

For non-volume observations, VOL alone is never sufficient for PASS. \\

Evidence constraints: \\
- If you mark any rule as satisfied, you MUST cite numeric values that support it. \\
- Each primary\_evidence item MUST include evidence\_json.bins and matching-length numeric arrays. \\
- If evidence does not support a satisfied check, verdict MUST be FAIL. \\

OUTPUT CONSTRAINT: \\
- Return EXACTLY ONE tool call to `distribution\_judgment\_tool`. \\
- Do NOT output any text outside the tool call. \\

DISTRIBUTION\_\allowbreak JUDGMENT\_\allowbreak USER\_\allowbreak TEMPLATE =\\

Formula: \\
- name: \{formula\_name\} \\
- definition: \{definition\} \\
- polarity: \{polarity\} \\

Observation: \\
- id: \{obs\_id\} \\
- description: \{obs\_description\} \\

EVIDENCE JSON: \\
\{evidence\_json\}
\end{promptbox}

\section{Run-to-Run Stability of LLM Validation}
\label{sec:secf}

To assess the robustness of LLM-based validation under stochastic decoding, we evaluate the variance of headline performance metrics across $10$ independent runs of the full pipeline (each run consisting of $3$ iterative rounds) on the CSI~500 universe with the same hypothesis prompt and identical thresholds. Table~\ref{tab:run_var} reports run-to-run variance. The small magnitudes indicate that validation outcomes are reproducible across independent runs and that the performance reported in Table~\ref{tab:tab1} is stable.

\begin{table}[ht]
\centering
\small
\renewcommand{\arraystretch}{1.15}
\begin{tabular*}{\columnwidth}{@{\extracolsep{\fill}}lc@{}}
\toprule
\textbf{Metric} & \textbf{Variance} \\
\midrule
Information Ratio (IR)        & 0.152 \\
Annualized Return             & 0.042 \\
Maximum Drawdown (MDD)        & 0.120 \\
Turnover                      & 0.046 \\
\bottomrule
\end{tabular*}
\caption{Run-to-run variance of FaVOR over $10$ independent runs.}
\label{tab:run_var}
\end{table}

\section{Strategy-Level Statistics}
\label{sec:secg}

Table~\ref{tab:trade_stats} reports trade-level distributional statistics from the CSI~500 out-of-sample period, verifying that FaVOR's cumulative excess return is not driven by a small number of extreme trades. The aggregated numbers confirm that gains are broadly distributed across many trades rather than concentrated in tail events.

\begin{table}[ht]
\centering
\small
\renewcommand{\arraystretch}{1.15}
\begin{tabular*}{\columnwidth}{@{\extracolsep{\fill}}lc@{}}
\toprule
\textbf{Metric} & \textbf{Value} \\
\midrule
Number of trades                  & 1,255 \\
Average holding period (days)     & 9.64 \\
Mean per-trade return (excess)            & 2.279\% \\
Profit factor                     & 1.986 \\
10th percentile of trade returns (excess)  & -7.24\% \\
90th percentile of trade returns (excess) & +8.52\% \\
\bottomrule
\end{tabular*}
\caption{Trade-level distributional statistics on out-of-sample period.}
\label{tab:trade_stats}
\end{table}

\section{Market-Regime Context for OOS 2025}
\label{sec:sech}

\begin{figure}[t]
    \centering
    \includegraphics[width=\columnwidth]{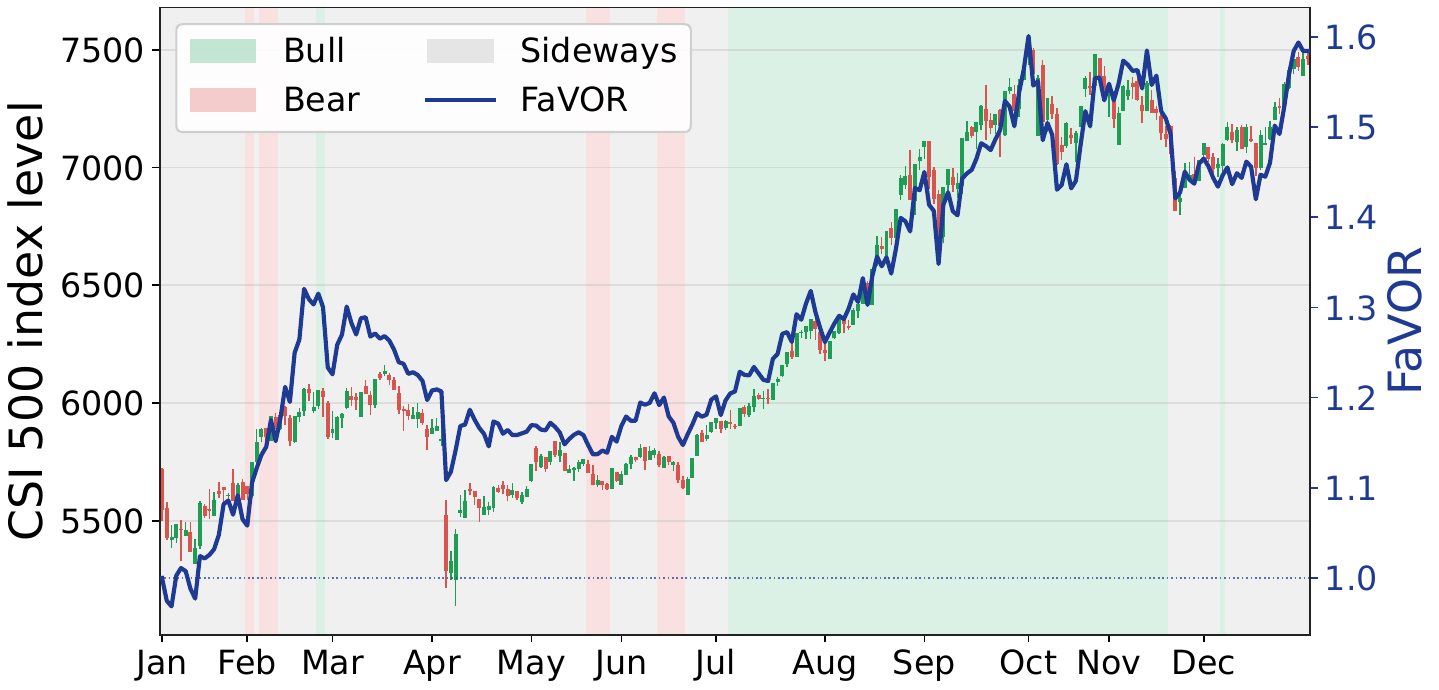}\\[1.2em]
    \includegraphics[width=\columnwidth]{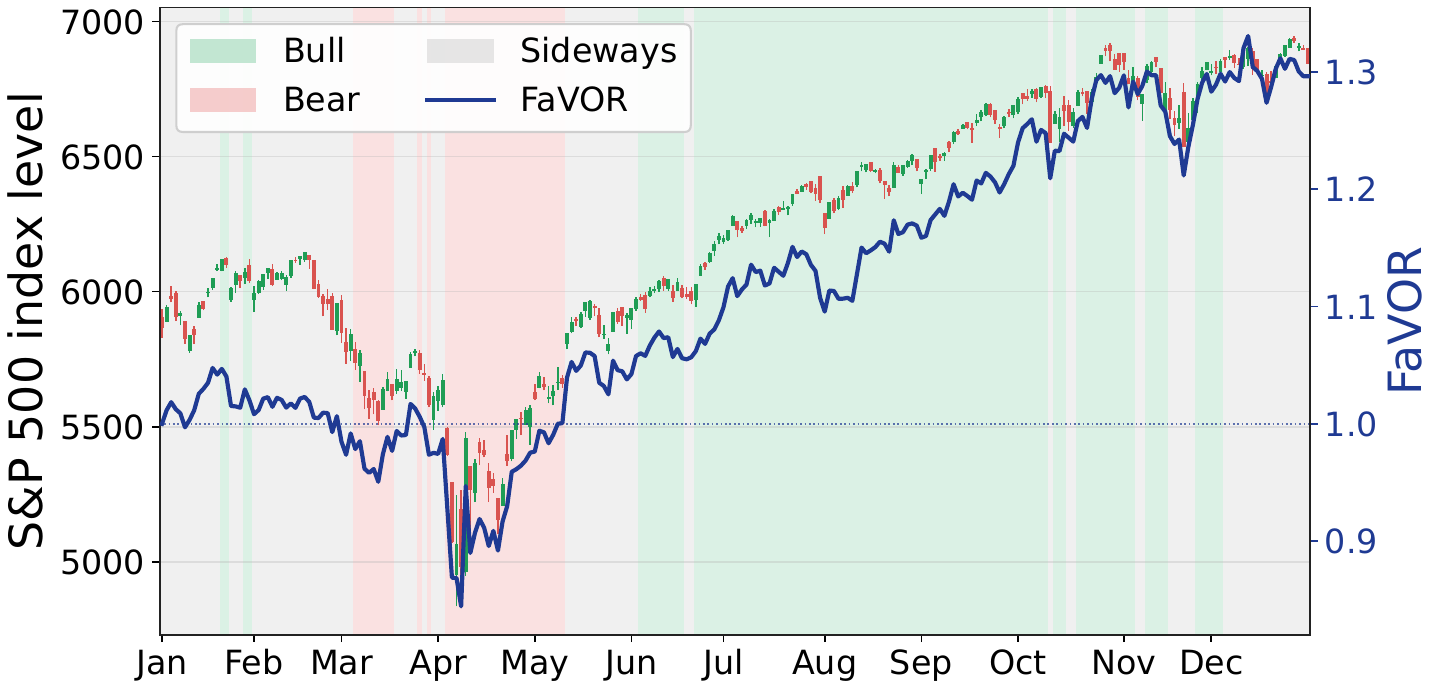}
    \caption{Out-of-sample 2025 market regimes for CSI~500 (top) and
    S\&P~500 (bottom), based on trailing 60-day benchmark returns, with FaVOR's cumulative return on the right axis (base = 1.0)}
    \label{fig:regime_2025}
\end{figure}

We summarize the 2025 out-of-sample market context for the two evaluation universes. In Figure~\ref{fig:regime_2025}, each trading day is labeled using the benchmark's trailing 60-day cumulative return: bull if $r_{60}{>}{+}5\%$, bear if $r_{60}{<}{-}5\%$, and sideways otherwise. These labels are used only for post-hoc interpretation and are not used by FaVOR during selection or trading.

\begin{table}[ht]
\centering
\small
\setlength{\tabcolsep}{4pt}
\begin{tabular*}{\columnwidth}{@{\extracolsep{\fill}}llc@{}}
\toprule
 & Formula & Pol. \\
\midrule
\multicolumn{3}{l}{\emph{CSI~500: 4 formulas, all positive polarity}} \\
$f_{006}$ & $\mathrm{SMA}(\mathrm{vol},7)\,/\,\mathrm{SMA}(\mathrm{vol},60)$        & $+$ \\
$f_{008}$ & $(\mathrm{close}-\mathrm{MA}_{30})\,/\,\mathrm{STD}_{30}(\mathrm{close})$ & $+$ \\
$f_{002}$ & $Z_{20}\!\bigl(\min_{10}\mathrm{low}\bigr)-Z_{20}\!\bigl(\max_{10}\mathrm{high}\bigr)$ & $+$ \\
$f_{010}$ & $\overline{\Delta\mathrm{vol}}_{10}$                                     & $+$ \\
\midrule
\multicolumn{3}{l}{\emph{S\&P~500: 3 formulas, mixed polarity}} \\
$f_{004}$ & $\mathrm{STD}_{7}(\mathrm{high}-\mathrm{low})$ & $+$ \\
$f_{001}$ & $\min_{20}\mathrm{close}\,/\,\max_{15}\mathrm{close}$ & $-$ \\
$f_{007}$ & $\mathrm{STD}_{15}(\mathrm{close})$ & $+$ \\
\bottomrule
\end{tabular*}
\caption{Formula combinations selected for the two out-of-sample portfolios.}
\label{tab:selected_formulas}
\end{table}

The CSI~500 window is mostly a long consolidation followed by a second-half rally. The index traded in a narrow range from January to early July, with only brief bear intervals, and then entered a sustained bull phase into year-end. The S\&P~500 window contains a sharper stress episode: the index fell from its February peak to an April low before recovering through the rest of the year. Thus, the two test universes expose FaVOR to different out-of-sample market paths: a consolidation-to-rally path in CSI~500 and a drawdown-to-recovery path in S\&P~500.

For completeness, Table~\ref{tab:selected_formulas} reports the formula combinations selected for the two out-of-sample portfolios. The table is included only to identify the selected signals.

\section{Hypothesis \& Formula Consistency Filter}
\label{sec:review_stage2}

Stage~2 is intended to test whether a candidate formula measures the OHLCV-observable condition stated in its originating hypothesis. It is not designed as a standalone alpha, IC, or RankIC filter. We therefore examine its strategy-level contribution separately from the marginal predictive association of individual formulas.

\paragraph{Stage-removal ablation on both markets.} Removing Stage~2 allows formulas that are not validated against their intended observable condition to enter factor integration. Table~\ref{tab:review_stage2_ablation} shows that the resulting excess IR deteriorates in both markets.

\begin{table}[ht]
\centering
\small
\renewcommand{\arraystretch}{1.12}
\begin{tabular*}{\columnwidth}{@{\extracolsep{\fill}}lcc@{}}
\toprule
\textbf{Market} & \textbf{Ours} & \textbf{W/O Stage~2} \\
\midrule
CSI~500 & 1.5295 & 0.1526 \\
S\&P~500 & 1.1315 & -0.8831 \\
\bottomrule
\end{tabular*}
\caption{Effect of removing Stage~2 factor-level validation. All settings
other than the presence of Stage~2 are held fixed.}
\label{tab:review_stage2_ablation}
\end{table}

\paragraph{PASS \& FAIL formulas.} We further evaluate first-iteration formulas before outer-loop feedback: 437 Stage~2 PASS formulas and 24 FAIL formulas. For each formula, we compute daily cross-sectional RankIC with forward returns at four horizons. As shown in Table~\ref{tab:review_pass_fail_rankic}, PASS formulas do not exhibit a uniform standalone RankIC advantage. This result is consistent with the role of Stage~2 as a semantic and distributional consistency filter rather than an individual-factor return predictor.

\begin{table}[ht]
\centering
\small
\renewcommand{\arraystretch}{1.10}
\begin{tabular*}{\columnwidth}{@{\extracolsep{\fill}}llcc@{}}
\toprule
\textbf{Market} & \textbf{Horizon} & \textbf{PASS RankIC} & \textbf{FAIL RankIC} \\
\midrule
CSI~500 & 1 day  & +0.0039 & +0.0025 \\
CSI~500 & 3 days & +0.0025 & +0.0030 \\
CSI~500 & 5 days & +0.0023 & +0.0035 \\
CSI~500 & 10 days & +0.0023 & +0.0047 \\
\midrule
S\&P~500 & 1 day  & +0.0040 & +0.0062 \\
S\&P~500 & 3 days & +0.0027 & +0.0073 \\
S\&P~500 & 5 days & +0.0017 & +0.0065 \\
S\&P~500 & 10 days & +0.0021 & +0.0057 \\
\bottomrule
\end{tabular*}
\caption{Mean single factor RankIC for Stage~2 PASS and FAIL formulas.
Formula values are oriented so that larger values correspond to a stronger
stated observation.}
\label{tab:review_pass_fail_rankic}
\end{table}

A positive forward return association does not imply that a formula measures the state described by its observation. For example, the candidate \texttt{-TS\_SUM(DELTA(low,1),5)} has positive RankIC on both markets across all examined horizons, but larger values correspond to continuing new lows rather than weakening downside pressure. Stage~2 rejects this candidate because its induced OHLCV distribution contradicts the stated condition. Thus, the filter removes formulas that may be predictive while representing the opposite market state from the hypothesis.

\section{Auditability and Reliability of Semantic Validation}
\label{sec:review_reliability}

The validation procedure separates deterministic evidence construction from a rule constrained semantic judgment. The deterministic component forms factor quantile bins and computes OHLCV statistics, including location, tail, and multi statistic evidence. The validation agent then assesses whether this evidence agrees with the textual observation and its polarity. The final decision is therefore auditable as a conjunction of explicit statistical checks and semantic consistency, rather than as an unconstrained language model judgment.

\begin{table}[ht]
\centering
\scriptsize
\renewcommand{\arraystretch}{1.10}
\setlength{\tabcolsep}{3pt}
\begin{tabular*}{\columnwidth}{@{\extracolsep{\fill}}lp{0.54\columnwidth}r@{}}
\toprule
\textbf{Audit} & \textbf{Design} & \textbf{Cases} \\
\midrule
Coded-rule comparison &
Validator verdicts are compared with an independent implementation of the same
location, tail, multi statistic, and contradiction checklist. &
8,362 \\
Expert assessment &
Experts receive the observation, formula, polarity, and induced distributional
evidence; verdict, model identity, and returns are hidden. &
100 \\
\bottomrule
\end{tabular*}
\caption{Reliability study design for semantic validation.}
\label{tab:review_reliability_design}
\end{table}

The blinded set contains 100 cases (52 CSI~500 and 48 S\&P~500), balanced between 50 framework PASS and 50 framework FAIL decisions from 44 observation groups; both audits assess semantic agreement rather than portfolio return.

The reliability analyses target the semantic judgment itself rather than portfolio return. They test whether the validator's interpretation of a fixed packet is aligned with an independently coded checklist and with blinded expert assessment.

The blinded set contains 100 cases (52 CSI~500 and 48 S\&P~500), balanced between 50 framework PASS and 50 framework FAIL decisions from 44 observation groups. Both audits assess semantic agreement rather than portfolio return.

\paragraph{Expert assessment outcomes.} Table~\ref{tab:review_expert_assessment} summarizes the expert assessment of semantic alignment for the blinded evidence packets. Experts judged 39 cases as agreement, 49 cases as partial agreement, and 12 cases as disagreement.

\begin{table}[ht]
\centering
\small
\renewcommand{\arraystretch}{1.10}
\begin{tabular*}{\columnwidth}{@{\extracolsep{\fill}}lr@{}}
\toprule
\textbf{Assessment} & \textbf{Cases} \\
\midrule
Disagreement & 12 \\
Partial agreement & 49 \\
Agreement & 39 \\
\bottomrule
\end{tabular*}
\caption{\footnotesize Expert assessment of semantic alignment in the blinded
evaluation set.}
\label{tab:review_expert_assessment}
\end{table}

Partial agreement is the most frequent outcome. In these cases, experts generally agreed with the observed antecedent condition and the stated directional outcome, but judged that the intermediate market mechanism required further discussion or more explicit justification. Thus, the evidence was consistent with the stated observation at the level of observable market behavior, while the mechanism connecting the condition to the expected outcome was not always fully established.

These results support the use of the validator as an auditable semantic consistency check. They do not imply that the validation procedure establishes a causal or fundamental economic mechanism. Instead, FaVOR evaluates whether a formula and its factor conditioned OHLCV evidence remain consistent with the observable condition stated in the hypothesis.

\section{Longitudinal Evaluation Across Test Years}
\label{sec:review_longitudinal}

To examine whether a fixed formula combination remains useful outside a single calendar boundary, we select one combination per market using a 2015~2019 training period and a 2020 validation period. The selected formula set and quantile level are held fixed for all subsequent tests; only per ticker cutoffs are estimated from the immediately preceding validation window and then frozen during each test period.

Table~\ref{tab:review_yearly_fixed_combo} reports annualized excess return and excess IR across five test years. The evaluation covers the 2022 market decline as well as subsequent market environments.

\begin{table}[ht]
\centering
\small
\renewcommand{\arraystretch}{1.10}
\resizebox{\columnwidth}{!}{%
\begin{tabular}{lrrrr}
\toprule
\textbf{Test year} &
\textbf{CSI~500 AR} & \textbf{CSI~500 IR} &
\textbf{S\&P~500 AR} & \textbf{S\&P~500 IR} \\
\midrule
2021 & +0.6964 & +1.7117 & +0.0564 & +0.4356 \\
2022 & +0.1356 & +0.4543 & +0.2377 & +0.9225 \\
2023 & +0.3317 & +1.2020 & +0.3590 & +2.5526 \\
2024 & +0.0449 & +0.1259 & +0.3340 & +1.2453 \\
2025 & +0.1310 & +0.5819 & +0.0826 & +0.5362 \\
\bottomrule
\end{tabular}%
}
\caption{Per year performance of the fixed combination selected on validation data.
The combination is selected before all reported test years. For each test,
cutoffs are estimated from the preceding validation window and held fixed
throughout the test period.}
\label{tab:review_yearly_fixed_combo}
\end{table}

We also evaluate the fixed combinations over 49 monthly start, 12 month test windows. Adjacent windows overlap and are therefore interpreted as start date sensitivity checks rather than independent trials.

\begin{table}[ht]
\centering
\small
\renewcommand{\arraystretch}{1.10}
\resizebox{\columnwidth}{!}{%
\begin{tabular}{lrrrrr}
\toprule
\textbf{Market} & \textbf{CR $>$ 0} & \textbf{CR} &
\textbf{IR} & \textbf{AR} & \textbf{MDD} \\
\midrule
CSI~500 & 33 / 49 & +0.1931 & +0.5919 & +0.1894 & -0.2649 \\
S\&P~500 & 39 / 49 & +0.1964 & +0.9119 & +0.1866 & -0.1406 \\
\bottomrule
\end{tabular}%
}
\caption{Sensitivity of fixed validation selected combinations.
Each result uses a trailing 24 month validation window and a following
12 month test window. Each metrics are arithmetic means over all
49 start test windows.}
\label{tab:review_monthly_fixed_combo}
\end{table}

\section{Hypothesis Provenance and Failure Analysis}
\label{sec:review_provenance}

FaVOR is not driven by a single manually selected hypothesis. Across saved runs, ten seed concepts generated 520 unique hypotheses over 696 outer-loop iterations. Table~\ref{tab:review_funnel} summarizes the progression from hypothesis generation to positive out-of-sample combinations.

\begin{table}[ht]
\centering
\small
\renewcommand{\arraystretch}{1.10}
\resizebox{\columnwidth}{!}{%
\begin{tabular}{lrrr}
\toprule
\textbf{Stage} & \textbf{All runs} & \textbf{Paper split} &
\textbf{Sell-off hypothesis} \\
\midrule
Distinct seed concepts & 10 & 9 & 1 \\
Unique LLM hypotheses & 520 & 486 & 67 \\
Outer-loop hypothesis iterations & 696 & 622 & 81 \\
Candidate formulas generated & 6,782 & 6,058 & 716 \\
Stage~3 candidate combinations & 18,699 & 16,741 & 1,148 \\
Iterations passing Stage~3 & 49 & 41 & 5 \\
Combinations with positive OOS return & 1,118 & 832 & 47 \\
\bottomrule
\end{tabular}%
}
\caption{Hypothesis to strategy audit across saved runs. The sell-off row
corresponds to the hypothesis used for the illustrative factor-validation case
study.}
\label{tab:review_funnel}
\end{table}

Only 49 of 696 hypothesis iterations clear the Stage~3 gate. Failure is most common for mechanisms that daily OHLCV data cannot cleanly identify, including volume dominant, short covering, and liquidity driven states. This result clarifies the scope of the framework: FaVOR is designed to enforce hypothesis formula consistency for OHLCV observable conditions, not to recover every economically plausible trading mechanism.

\section{Sensitivity to Selectivity and Directional Scope}
\label{sec:review_selectivity}

We assess how global selectivity affects the universe of Stage~3 combinations. Table~\ref{tab:review_sigma_sensitivity} reports the unconditional mean 2025 cumulative excess return over all 649 Stage~3 combinations under a common global selectivity level. The main method instead selects formula specific thresholds on validation data.

\begin{table}[ht]
\centering
\small
\renewcommand{\arraystretch}{1.12}
\begin{tabular*}{\columnwidth}{@{\extracolsep{\fill}}lcc@{}}
\toprule
\textbf{Global selectivity $\sigma$} & \textbf{CSI~500 CR} & \textbf{S\&P~500 CR} \\
\midrule
0.55 & +0.2048 & +0.0817 \\
0.75 & +0.0958 & +0.0340 \\
0.90 & +0.0012 & +0.0245 \\
0.95 & +0.0048 & +0.0134 \\
\bottomrule
\end{tabular*}
\caption{Unconditional mean 2025 cumulative excess return across all
Stage~3 combinations under a shared global selectivity level.}
\label{tab:review_sigma_sensitivity}
\end{table}

Directional Selectivity is intentionally not a universal profitability filter. Among 212 evaluated Stage~3 combinations, 81 were rejected because their realized direction did not remain stable as selectivity tightened. Of these rejected combinations, 47 still obtained positive after cost cumulative return in the held out evaluation. The filter therefore discards a meaningful set of profitable but directionally unstable signals, especially rebound and panic selling patterns. This exclusion is the cost of restricting FaVOR to directionally consistent signals.

\section{Variance and Baseline Context}
\label{sec:review_variance}

Table~\ref{tab:review_variance} reports sample variances and their corresponding standard deviations. FaVOR runs use the same high level hypothesis prompt and threshold configuration to isolate stochastic decoding effects, whereas RD-Agent results are based on ten independent final outputs under the common 2022 / 2023 / 2024 / 2025 split.

\begin{table}[ht]
\centering
\small
\renewcommand{\arraystretch}{1.12}
\begin{tabular*}{\columnwidth}{@{\extracolsep{\fill}}llrr@{}}
\toprule
\textbf{Method} & \textbf{Metric} & \textbf{Variance} & \textbf{Std.} \\
\midrule
FaVOR & Information Ratio & 0.152 & 0.390 \\
FaVOR & Annualized Return & 0.042 & 0.205 \\
FaVOR & Maximum Drawdown & 0.120 & 0.346 \\
\midrule
RD-Agent & Information Ratio & 0.191 & 0.438 \\
RD-Agent & Annualized Return & 0.033 & 0.182 \\
RD-Agent & Maximum Drawdown & 0.055 & 0.235 \\
\bottomrule
\end{tabular*}
\caption{Run level dispersion of FaVOR and RD-Agent. The table describes
observed run to run variation and is not used to claim a statistically
significant return gap.}
\label{tab:review_variance}
\end{table}

All compared methods use the same OHLCV derived inputs, data splits, transaction costs, and backtest environment. Their inductive priors and search capacities nevertheless differ: FaVOR uses language model priors to propose hypotheses and formulas, while ML/DL baselines use the fixed Alpha158 feature family and RL baselines search OHLCV primitives without a natural language hypothesis prior. We therefore interpret the benchmark comparison as a comparison under matched data and execution conditions, rather than as a claim of matched search budgets.

\section{Scope of the Interpretability Claim}
\label{sec:review_scope}

FaVOR provides structural traceability from a natural language hypothesis to observable conditions, formulas, and factor conditioned evidence. Its interpretability claim is therefore limited to empirical consistency between a formula and the OHLCV observable condition that motivated it. The framework does not establish a causal or fundamental economic mechanism, and this distinction is important when interpreting its validation decisions and portfolio results.

\section{Computational Cost}
\label{sec:secj}

All local experiments were run on a single workstation equipped with two NVIDIA RTX PRO 6000 Blackwell Server Edition GPUs (96GB of memory each), two 16-core Intel64 CPUs (32 physical cores and 64 logical threads in total, dual-socket).

Closed-source backbones were accessed through commercial API endpoints; their parameter counts are not publicly disclosed by the providers. Open-weight backbones were self-hosted on the local GPUs using Ollama. Llama-3.3-70B is a dense 70B-parameter model, and Qwen3-235B is a Mixture-of-Experts model with 235B total parameters and 22B active parameters per token. For API-based runs, we report token counts and monetary cost because provider-side GPU hours are not observable. For self-hosted open-weight runs, inference ran on the two local GPUs described above.

Tables~\ref{tab:llm_usage_csi} and~\ref{tab:llm_usage_sp} summarize LLM usage per call by market. Table~\ref{tab:llm_usage_stage} gives an example per-stage breakdown for one GPT-4o run on CSI~500. Token counts include both prompt and completion tokens, and API costs are estimated from the provider pricing used during the experiments.

\begin{center}
\small
\renewcommand{\arraystretch}{1.15}
\begin{tabular*}{\columnwidth}{@{\extracolsep{\fill}}lrr@{}}
\toprule
Backbone & Avg tokens/call & Avg cost/call \\
\midrule
GPT-4o            & 4{,}287 & \$0.01540 \\
GPT-5.4-mini      & 4{,}097 & \$0.00090 \\
Gemini-2.5-flash  & 5{,}525 & \$0.00119 \\
Claude-Sonnet-4.6 & 7{,}732 & \$0.00198 \\
\bottomrule
\end{tabular*}
\captionof{table}{LLM usage per call on CSI~500. Token counts include both prompt
and completion tokens.}
\label{tab:llm_usage_csi}
\end{center}

\begin{center}
\small
\renewcommand{\arraystretch}{1.15}
\begin{tabular*}{\columnwidth}{@{\extracolsep{\fill}}lrr@{}}
\toprule
Backbone & Avg tokens/call & Avg cost/call \\
\midrule
GPT-4o            & 4{,}186 & \$0.01478 \\
GPT-5.4-mini      & 4{,}171 & \$0.00095 \\
Gemini-2.5-flash  & 5{,}335 & \$0.00118 \\
Claude-Sonnet-4.6 & 7{,}421 & \$0.00186 \\
\bottomrule
\end{tabular*}
\captionof{table}{LLM usage per call on S\&P~500. Token counts include both prompt
and completion tokens.}
\label{tab:llm_usage_sp}
\end{center}

\begin{center}
\small
\renewcommand{\arraystretch}{1.15}
\begin{tabular*}{\columnwidth}{@{\extracolsep{\fill}}lrr@{}}
\toprule
Stage & Avg tokens/call & Avg cost/call \\
\midrule
Hypothesis       & 1{,}400 & \$0.00433 \\
Observation        & 1{,}425 & \$0.00487 \\
Factor generation          & 4{,}525 & \$0.01904 \\
Self-correction     & 5{,}193 & \$0.02211 \\
Refinement          & 8{,}435 & \$0.02983 \\
Validation & 3{,}996 & \$0.01295 \\
\bottomrule
\end{tabular*}
\captionof{table}{Example LLM usage for one gpt-4o run on CSI~500.
Reported per call within each stage; token counts include both prompt
and completion tokens.}
\label{tab:llm_usage_stage}
\end{center}

\section{Software, Models, and Data}
\label{sec:seck}

Our backtest pipeline builds on Qlib \cite{yang2020qlib} for portfolio simulation, transaction-cost accounting, and benchmark excess-return evaluation, and on Optuna \cite{10.1145/3292500.3330701} for Stage~4 threshold search on the validation split. Qlib, Optuna, LightGBM \cite{ke2017lightgbm}, AlphaAgent \cite{tang2025alphaagent}, R\&D-Agent-Quant \cite{li2025r}, AlphaQCM \cite{zhu2025alphaqcm}, and Ollama are released under the MIT license. XGBoost \cite{chen2016xgboost} is released under Apache-2.0, and scikit-learn \cite{pedregosa2011scikit} is released under the BSD 3-Clause license. AlphaForge \cite{shi2025alphaforge} is released as an open-source research codebase without a formal license, and we use it solely for non-commercial research consistent with its stated purpose.

Open-weight backbones Llama-3.3-70B~\cite{grattafiori2024llama} and Qwen3-235B~\cite{yang2025qwen3} are released under the Llama 3.3 Community License and Apache-2.0, respectively. Closed-source backbones GPT-4o \cite{hurst2024gpt}, GPT-5.4-mini \cite{openai2026gpt54thinking}, Gemini-2.5-Flash \cite{comanici2025gemini}, and Claude-Sonnet-4.6 \cite{anthropic2026claudesonnet46} are accessed through the OpenAI, Google, and Anthropic APIs under each provider's terms of use.

CSI~500 price and volume data are obtained via the \texttt{baostock} Python package, whose package metadata lists a BSD license. S\&P~500 price and volume data are obtained via the \texttt{yfinance} Python package, whose package metadata lists the Apache Software License. The underlying Yahoo Finance data are subject to Yahoo's terms of use and applicable data-provider restrictions; we use them only for non-commercial academic evaluation, report only derived experimental results, and do not redistribute raw market data.

\end{document}